\documentclass[10pt,twocolumn,letterpaper]{article}

\usepackage[pagenumbers]{wacv} 

\definecolor{wacvblue}{rgb}{0.21,0.49,0.74}
\usepackage[pagebackref,breaklinks,colorlinks,allcolors=wacvblue]{hyperref}
\usepackage{hyperref}
\usepackage{pifont}
\usepackage{xcolor}

\newcommand{\resubmit}[1]{\textcolor{black}{#1}}

\def\wacvPaperID{868} 
\def\confName{WACV}
\def\confYear{2027}

\title{FlexTry-On: A Time-Efficient Framework for Multi-Scenario Virtual Try-on}
\author{
Phuong Dam \and
Quyen Nguyen Duc \and
Son Hoang Minh \and
Daeyoung Kim\\
Korea Advanced Institute of Science and Technology (KAIST)\\
{\tt\small \{hoangphuong1211,nguyenducquyen2311,sonhm2910,kimd\}@kaist.ac.kr}
}

\begin{document}
\twocolumn[{%
\renewcommand\twocolumn[1][]{#1}%
\maketitle
\begin{center}
    \centering
    \scalebox{1.0}{%
    \includegraphics[width=\textwidth]{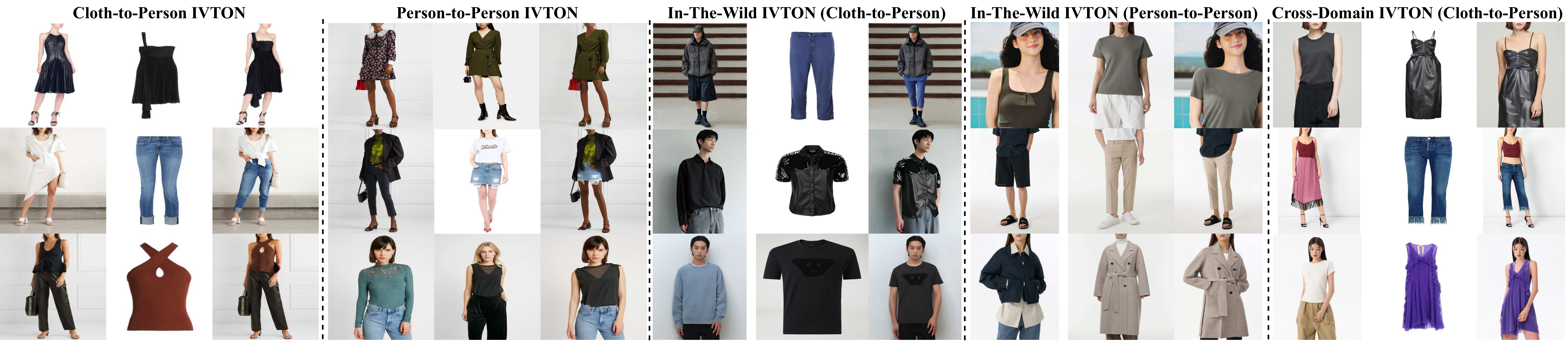}}
    \vspace{-6mm}
    \captionsetup{labelfont=scriptsize,textfont=scriptsize}
    \captionof{figure}{Qualitative results of \textbf{FlexTry-On} across multiple IVTON scenarios, including Cloth-to-Person, Person-to-Person on the DressCode dataset, In-the-Wild settings, and Cross-Domain IVTON. Please zoom in for a clearer visualization of fine details.}
    \label{fig:teaser}
\end{center}
}]


\begin{abstract}
Virtual try-on has achieved remarkable progress in recent years, yet most existing methods are designed for a specific scenario, such as cloth-to-person or person-to-person try-on, and often require separate conditioning strategies and processing pipelines for different settings. We present FlexTry-On, a unified and time-efficient diffusion-based framework for diverse virtual try-on scenarios, including cloth-to-person, person-to-person, cross-domain, and in-the-wild settings. FlexTry-On adopts a common conditioning formulation, where a text-aware condition extraction strategy converts heterogeneous garment sources into a unified representation, and a dual-conditioning diffusion transformer jointly models garment appearance and person identity. Extensive experiments demonstrate competitive performance on both cloth-to-person and person-to-person benchmarks while maintaining fast inference. Moreover, the same framework can be directly applied to unseen cross-domain scenarios without requiring separate architectures or scenario-specific processing pipelines. These results highlight the potential of unified and efficient virtual try-on frameworks as a practical step toward more flexible and widely applicable virtual try-on systems.

\end{abstract}

\vspace{-5mm}
\section{Introduction}
\label{sec:intro}

Image-based virtual try-on (IVTON) aims to synthesize realistic images of a target person wearing a desired garment while preserving both garment appearance and human identity~\cite{cordier20012d}. Recent advances in diffusion models~\cite{ho2020denoising} have significantly improved virtual try-on quality, leading to highly realistic results on standard benchmarks~\cite{morelli2023ladi,choi2024improving,dam2024time,hong2025ita,kim2025promptdresser}. However, most existing methods are developed for a specific setting, such as cloth-to-person (C2P) or person-to-person (P2P) virtual try-on. In practice, garments may originate from clean product images, reference persons, or unconstrained real-world photographs. Existing virtual try-on systems, however, are typically tailored to a particular input format and scenario. Extending them to cross-domain or in-the-wild settings often requires additional assumptions, preprocessing steps, or task-specific adaptations, which also introduce additional computational overhead. As a result, virtual try-on remains fragmented across scenarios despite sharing the same underlying objective of garment transfer. A key challenge is how to convert garments from heterogeneous sources into a unified conditioning representation while preserving garment appearance and person identity.

\begin{center}
\emph{Can a single virtual try-on framework effectively handle diverse virtual try-on scenarios?}
\end{center}



In this paper, we present \textbf{FlexTry-On}, a unified and time-efficient diffusion-based framework for cloth-to-person, person-to-person, cross-domain, and in-the-wild virtual try-on. To accommodate heterogeneous garment sources, FlexTry-On first employs a Text-Aware Segmentation Module (TASM) that extracts task-aware garment conditions from either product images or reference persons. These conditions are converted into a unified conditioning representation and processed by a Dual-DiT Tryon Module (DDTM), where a reference branch captures garment appearance and a try-on branch synthesizes the final image while preserving person identity. In addition, we employ a Style-GRAM loss to encourage appearance consistency between the generated garment region and the reference garment, improving garment preservation under noisy supervision. Together, these components enable the same framework to support diverse virtual try-on scenarios through a shared conditioning and generation pipeline.

By formulating diverse virtual try-on tasks under a common conditional generation framework, FlexTry-On eliminates the need for scenario-specific architectures and maintains a consistent inference pipeline across different settings. Furthermore, the same framework supports multiple garment categories within a unified conditioning formulation, enabling a unified formulation across different virtual try-on scenarios.

\begin{figure}[t]

  \centering
  \includegraphics[width=1.0\linewidth]{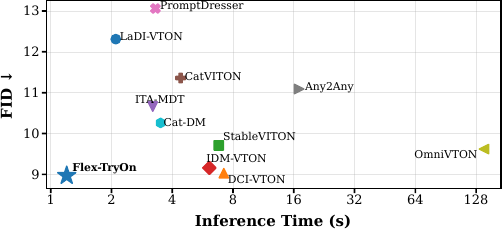}
  \vspace{-6mm}
  \captionsetup{labelfont=scriptsize,textfont=scriptsize}
  \caption{Comparison of FlexTry-On with other diffusion-based methods using the VITON-HD dataset (C2P). Lower values indicate better performance for both FID and inference speed. FlexTry-On supports all evaluated virtual try-on scenarios within a unified framework, as summarized in Tab.~\ref{tab:fullresult}.}
  \label{fig:intro-graph}
   \vspace{-6mm} 
\end{figure}

Extensive experiments demonstrate that FlexTry-On achieves competitive visual quality and strong FID \cite{heusel2017gans} performance on both C2P and P2P benchmarks while maintaining efficient inference. In addition, the same framework can be directly applied to challenging cross-domain and in-the-wild scenarios. As shown in Fig.~\ref{fig:intro-graph}, FlexTry-On achieves a favorable balance between synthesis quality, inference efficiency, and scenario flexibility, providing a practical step toward more flexible and widely applicable virtual try-on systems.
Our contributions are summarized as:

\begin{enumerate}
\item We propose FlexTry-On, a unified and time-efficient diffusion-based framework that bridges cloth-to-person, person-to-person, cross-domain, and in-the-wild virtual try-on scenarios within a single framework.


\item We introduce a unified virtual try-on framework consisting of a Text-Aware Segmentation Module (TASM) and a Dual-DiT Tryon Module (DDTM), enabling conditioning from heterogeneous garment sources while jointly modeling garment appearance and person identity. We further employ a Style-GRAM loss for garment appearance preservation under noisy supervision.

\item Extensive evaluations on both standard benchmarks and challenging real-world settings demonstrate competitive visual quality, strong FID performance, and efficient inference within a unified framework.

\end{enumerate}

\section{Related Works}
\label{sec:related_work}
IVTON~\cite{cordier20012d} has evolved from warping-based pipelines to diffusion-based generative frameworks. Early methods~\cite{han2018viton,raj2018swapnet,wang2018toward,choi2021viton,lee2022high,dam2024time} rely on explicit garment alignment and image synthesis, making them sensitive to pose variations, occlusions, and inaccurate geometric estimation. Recent diffusion-based approaches~\cite{zhu2023tryondiffusion,morelli2023ladi,gou2023taming,kim2024stableviton,choi2024improving,zeng2024cat,chong2024catvton,kim2025promptdresser,guo2025any2anytryon,lee2025voost,li2025dit} significantly improve visual quality through large pre-trained models and attention-based mechanisms.

Despite these advances, most existing methods are designed around specific conditioning assumptions. Many diffusion-based IVTON approaches assume a particular garment representation, such as clean product images, extracted garment regions, or reference-person-based garment cues, and design their conditioning mechanisms accordingly~\cite{zhu2023tryondiffusion,morelli2023ladi,gou2023taming,kim2024stableviton,choi2024improving,zeng2024cat,chong2024catvton,kim2025promptdresser,guo2025any2anytryon,lee2025voost,li2025dit}. Consequently, their conditioning mechanisms are often tailored to specific input formats, limiting both flexibility and efficiency across different virtual try-on scenarios.

Furthermore, most existing methods ~\cite{zhu2023tryondiffusion,morelli2023ladi,gou2023taming,kim2024stableviton,choi2024improving,zeng2024cat,chong2024catvton,kim2025promptdresser,lee2025voost,li2025dit} implicitly assume category-consistent garment transfer, where the source and target garments belong to the same clothing category (e.g., upper-to-upper or dress-to-dress). Under this assumption, editable regions can be determined using predefined masks, human parsing labels, or garment extraction pipelines. However, transfers across different garment categories introduce ambiguous editable regions that cannot be uniquely determined using conventional category-based masking strategies.

Recent methods, including Mask Does Not Matter~\cite{du2025mask}, All Parts Matter~\cite{du2025all}, OmniTry~\cite{feng2026omnitry}, OmniVTON~\cite{yang2025omnivton}, and M\&M VTO~\cite{zhu2024m}, extend virtual try-on beyond the conventional cloth-to-person setting toward more diverse and realistic applications. Nevertheless, they are primarily designed for specific input assumptions and do not explicitly address category-crossing transfers. These observations suggest the key challenge is not only generating realistic try-on images but also designing a framework that can accommodate heterogeneous garment sources, identify task-aware editable regions, and support diverse virtual try-on scenarios without sacrificing inference efficiency.

\begin{figure*}[t]
  \centering
   \includegraphics[width=1.0\linewidth]{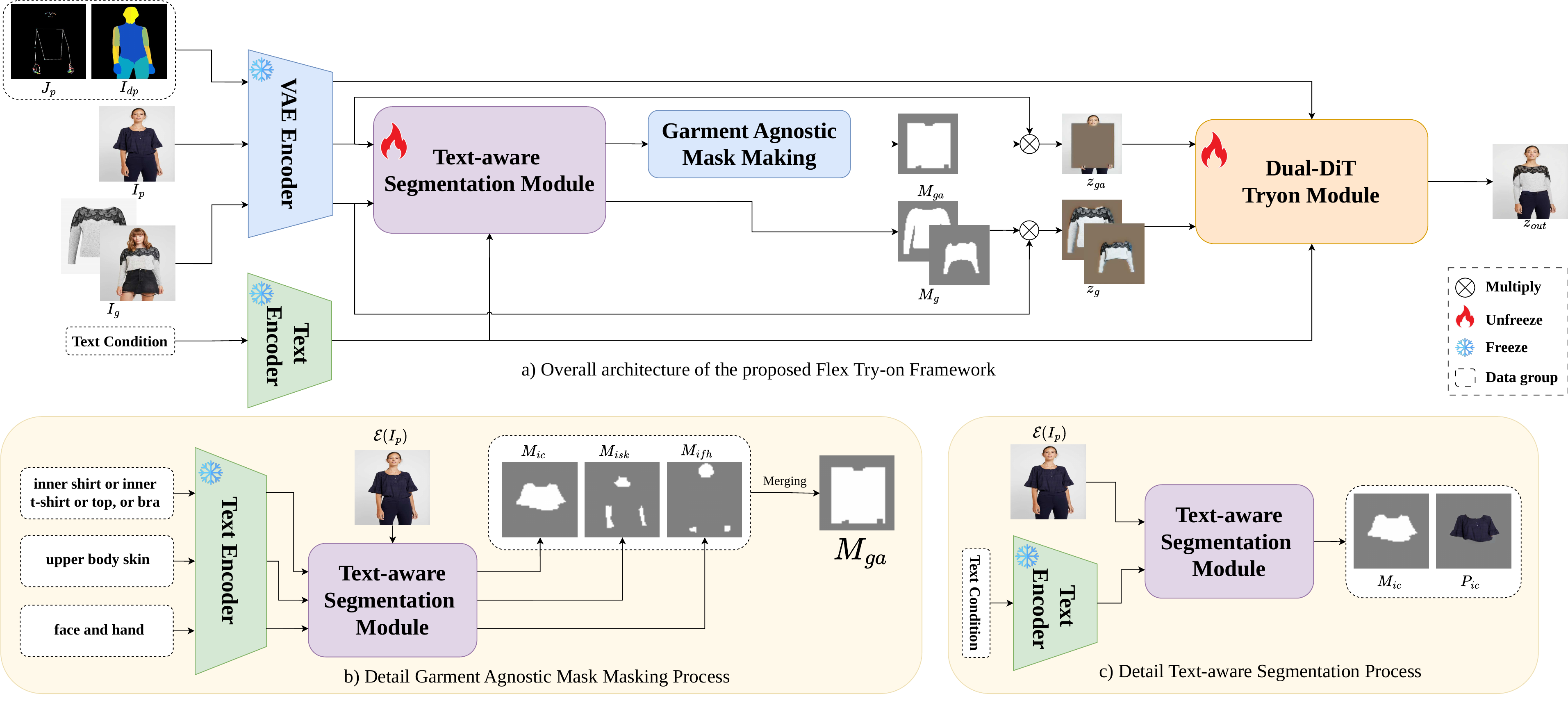}
   \vspace{-2mm}
   \captionsetup{labelfont=scriptsize,textfont=scriptsize}
   \caption{Overview of the FlexTry-On framework for IVTON. Multiple reference inputs, including pose keypoints \(J_p\) and DensePose maps \(I_{dp}\), are encoded into latent representations and concatenated as inputs to the diffusion process. The TASM module receives a latent image with text conditions embedded via the CLIP-Text Encoder~\cite {rombach2022high} from SD~2.0. The TASM outputs both the latent representations of the target RGB regions and their corresponding masks. In TASM, we illustrate the latent representation of the RGB input cloth $P_{ic}$ and its associated mask $M_{ic}$. The garment-agnostic mask \(M_{ga}\) is obtained by merging masks under different text conditions according to Eq.~\ref{eq:merge}}
   \label{fig:fullframework}
   \vspace{-6mm}
\end{figure*}

\section{Methodology}
\label{sec:method}
To support diverse virtual try-on scenarios within a single framework, we propose FlexTry-On, a diffusion-based approach that accommodates different input modalities and garment categories under a unified formulation. FlexTry-On consists of two key components: a Text-Aware Segmentation Module (TASM), which converts heterogeneous garment inputs into task-aware garment conditions and editable regions, and a Dual-DiT TryOn Module (DDTM), which synthesizes the final try-on image by jointly modeling garment appearance and person identity.

Built on a DiT-based latent diffusion framework~\cite{rombach2022high,peebles2023scalable}, FlexTry-On encodes inputs into a shared latent space using a pre-trained VAE. DDTM employs a dual-DiT design, where a reference DiT extracts garment representations only once and a try-on DiT performs image synthesis through denoising. This design enables a unified and computationally efficient pipeline for cloth-to-person, person-to-person, and cross-domain virtual try-on scenarios.




\begin{figure*}[t]
  \centering
   \includegraphics[width=1.0\linewidth]{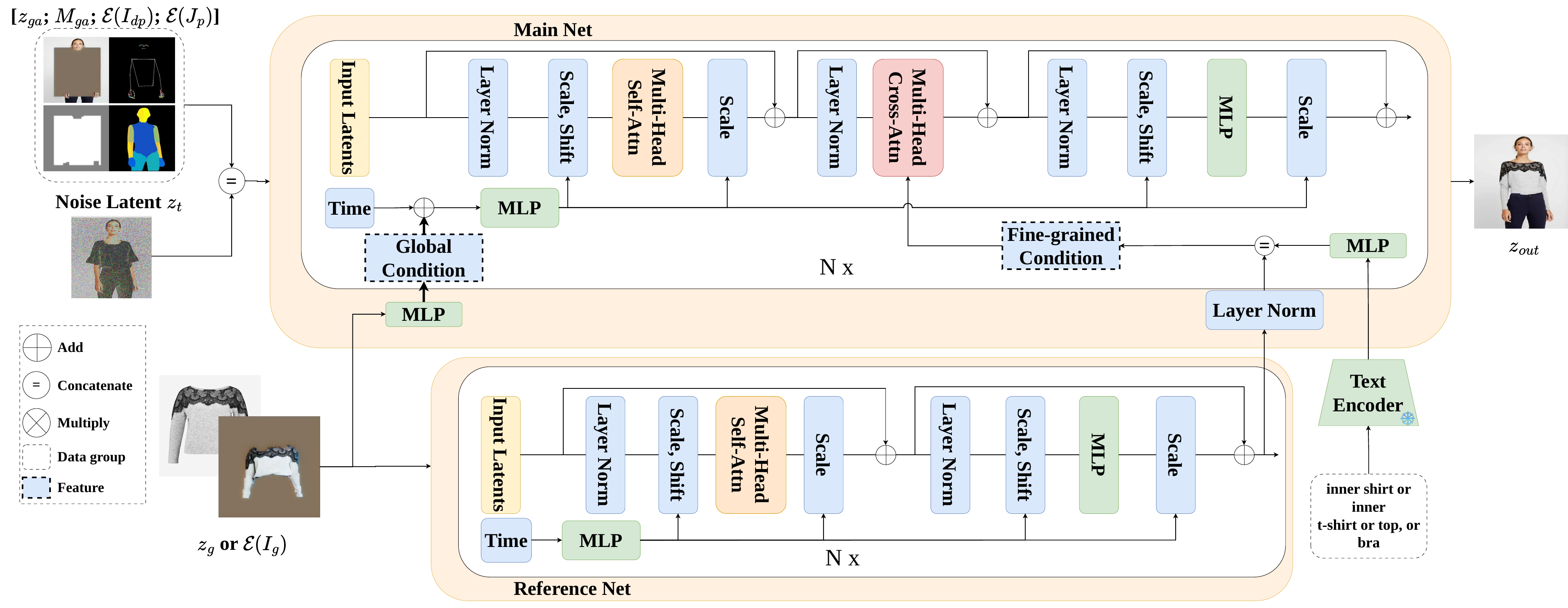}
    \captionsetup{labelfont=scriptsize,textfont=scriptsize}
   \caption{Overview of the Dual-DiT Tryon Module (DDTM). The DDTM follows a DiT-based architecture~\cite{peebles2023scalable} with cross-attention to integrate person and garment information. During training, different conditioning strategies are adopted for C2P and P2P settings. For clarity, patchification and positional embeddings are omitted.}
   \label{fig:ddtm}
   \vspace{-5mm}
\end{figure*}
\subsection{Text-Aware Segmentation Module (TASM)}
\label{sec: TASM}
To support diverse virtual try-on scenarios, we introduce a Text-Aware Segmentation Module (TASM) that converts heterogeneous inputs into a unified conditioning representation. Unlike conventional human parsing models~\cite{li2020self} or generic segmentation frameworks such as SAM~\cite{kirillov2023segment} and Grounded-SAM~\cite{liu2023grounding,ren2024grounded} that are designed for pixel-level segmentation, TASM is designed specifically for virtual
try-on conditioning and operates directly in the latent space. Given an image and predefined text conditions, TASM predicts garment-specific masks for downstream conditioning.

As illustrated in Fig.~\ref{fig:fullframework}, TASM adopts a DiT-based architecture~\cite{peebles2023scalable} and injects text conditions through multi-head cross-attention. Given a latent image representation, the module predicts region-specific masks corresponding to different garment categories. Specifically, only the predicted masks are retained for subsequent processing, while associated latent appearance features are discarded. \resubmit{An ablation of this design is provided in Appendix~\ref{sec:TASM_design_choice}}. The objective function is a combination of reconstruction \(\mathcal{L}_{1}, \mathcal{L}_{2}\) and adversarial objectives:
\begin{equation}
\mathcal{L}_{\text{textseg}} = \mathcal{L}_{1} + \mathcal{L}_{2} + \alpha_{\text{GAN}} \mathcal{L}_{\text{GAN}},
\end{equation}
where $\alpha_{\text{GAN}}$ controls the contribution of the adversarial loss~\cite{jolicoeur2018relativistic}. Ground-truth masks follow the protocol of~\cite{zhang2024mmtryon}, supplemented with additional annotations for facial, hair, and accessory regions from~\cite{li2020self}. 

To construct the garment-agnostic mask used in the try-on stage, TASM predicts region-specific masks according to predefined text conditions. These masks are merged as:
\begin{equation}
M_{ga} = \mathrm{RECT}\left[ M_{ic} \lor M_{isk} \right] \setminus M_{ifh},
\label{eq:merge}
\end{equation}
where $M_{ic}$, $M_{isk}$, and $M_{ifh}$ denote the clothing, skin, and face-with-hands masks, respectively, and $\mathrm{RECT}[\cdot]$ represents the enclosing rectangular region. This formulation enables flexible region control across cross-domain try-on settings (e.g., dress-to-upper or upper--lower-to-dress), as illustrated in Fig.~\ref{fig:cross-domain}. Unlike conventional human parsing \cite{li2020self, khirodkar2024sapiens}, TASM generates task-specific masks that explicitly identify the regions involved in the try-on process. By decoupling editable-region construction from predefined garment categories, TASM directly supports cross-domain virtual try-on scenarios where conventional category-dependent masking strategies become ambiguous. Additional discussion is provided in Appendix~\ref{sec:inpainting_mask_strategy}.

\subsection{Dual-DiT Tryon Module (DDTM)}

The Dual-DiT Tryon Module consists of two diffusion transformer
branches~\cite{peebles2023scalable}: a Reference Net and a Main Net, as
illustrated in Fig.~\ref{fig:ddtm}. The Reference Net extracts garment-related
features, while the Main Net performs try-on synthesis. This dual-stream design
decouples garment appearance modeling from the generation process, providing
structured guidance throughout denoising.

\resubmit{
Given a garment image $I_g$ and a person image $I_p$, TASM constructs the
garment condition
$z_g=\mathcal{E}(I_g)\odot M_g$, where $M_g$ denotes the garment mask.
Similarly, the garment-agnostic person condition is constructed as
$z_{ga}=\mathcal{E}(I_p)\odot M_{ga}$, where $M_{ga}$ is predicted by TASM.
For P2P, TASM extracts $M_g$ from the reference person to isolate the source
garment, whereas for C2P the clean product garment is directly retained.
Thus, the same interface accommodates both product-garment and
person-reference inputs.
}

\resubmit{
The Reference Net processes $z_g$ once at a fixed diffusion
timestep~\cite{choi2024improving}, producing multi-level garment features that
are reused throughout denoising. The Main Net receives the noisy latent $z_t$,
garment-agnostic latent $z_{ga}$, mask $M_{ga}$, pose keypoints
$\mathcal{E}(J_p)$, and DensePose map $\mathcal{E}(I_{dp})$. These spatial
conditions are patch-embedded and concatenated to form the global conditioning
input. Intermediate Reference-Net features are combined with garment-type text
embeddings and injected into the corresponding Main-Net layers through
cross-attention. This design provides multi-level garment guidance while
separating garment feature extraction from iterative image synthesis.
}

\subsubsection{Denoising Objective}
The training objective follows the standard latent diffusion formulation \cite{rombach2022high}:
\begin{equation}
\mathcal{L}_{denoise} =
\mathbb{E}_{y, c, \epsilon \sim \mathcal{N}(0, I),\, t}
\left[
\left\|
\epsilon - \epsilon_{\theta}(y_t, c, t)
\right\|^2
\right],
\label{eq:ldenoise}
\end{equation}
where \(y_t\) denotes the noisy latent at timestep \(t\), \(c\) represents the conditioning inputs including \(z_g\), \(z_{ga}\), \(M_{ga}\), \(\mathcal{E}(I_{dp})\), and \(\mathcal{E}(J_p)\), and \(\epsilon_{\theta}\) is the predicted noise.
\begin{table*}[t]
\centering
\caption{Quantitative comparison on VITON-HD and DressCode Upper in C2P.
Inf.~Time denotes method-specific inference time; shared pose/DensePose
preprocessing is excluded, with exceptions detailed in
Appendix~\ref{sec:imp_detail}. FlexTry-On is trained only on VITON-HD and
DressCode, whereas FlexTry-On* uses the full uncurated C2P--P2P corpus.
Best/second-best results are \textbf{bold}/\underline{underlined}.
Cross C2P/P2P denote cross-domain variants. DressCode Lower/Dresses results
are in Appendix~\ref{sec:quantitavedresscode}. $\dagger$: paper-reported.}
\label{tab:fullresult}

\resizebox{\textwidth}{!}{
\begin{tabular}{l|c|ccc|ccc|cccc}
\hline
\multicolumn{1}{c|}{} &
\multicolumn{1}{l|}{Inf. Time} &
\multicolumn{3}{c|}{VITON-HD} &
\multicolumn{3}{c|}{Dress Code - Upper} &
\multicolumn{4}{c}{Support IVTON Tasks} \\

\multicolumn{1}{c|}{Method} &
(s) $\downarrow$ &
FID (pair/unp.) $\downarrow$ &
SSIM $\uparrow$ &
LPIPS $\downarrow$ &
FID (pair/unp.) $\downarrow$ &
SSIM $\uparrow$ &
LPIPS $\downarrow$ &
C2P &
P2P &
Cross C2P &
Cross P2P \\

\hline

MDNM $\dagger$[IJCAI 2025] \cite{du2025mask}
& -
& -/9.38
& \underline{0.872}
& 0.086
& -
& -
& -
& \textcolor{green}{\ding{51}}
& \textcolor{red}{\ding{55}}
& \textcolor{red}{\ding{55}}
& \textcolor{red}{\ding{55}}
\\

APM $\dagger$[ICCV 2025] \cite{du2025all}
& -
& -/9.38
& \textbf{0.901}
& \textbf{0.079}
& -
& -
& -
& \textcolor{green}{\ding{51}}
& \textcolor{red}{\ding{55}}
& \textcolor{red}{\ding{55}}
& \textcolor{red}{\ding{55}}
\\

M\&MVTO [CVPR 2024] \cite{zhu2024m}
& -
& -
& -
& -
& -
& -
& -
& \textcolor{green}{\ding{51}}
& \textcolor{green}{\ding{51}}
& \textcolor{red}{\ding{55}}
& \textcolor{red}{\ding{55}}
\\

OmniVTON $\dagger$ [ICCV 2025] \cite{yang2025omnivton}
& 140
& 7.76/9.62
& 0.858
& 0.098
& -
& -
& -
& \textcolor{green}{\ding{51}}
& \textcolor{green}{\ding{51}}
& \textcolor{red}{\ding{55}}
& \textcolor{red}{\ding{55}}
\\

\hline

LaDI-VTON [ACM 2023]\cite{morelli2023ladi}
& \underline{2.1}
& 9.63/12.31
& 0.858
& 0.098
& 10.45/14.05
& 0.927
& 0.062
& \textcolor{green}{\ding{51}}
& \textcolor{red}{\ding{55}}
& \textcolor{red}{\ding{55}}
& \textcolor{red}{\ding{55}}
\\

DCI-VTON [ACM 2023]\cite{gou2023taming}
& 7.2
& \underline{5.56}/\underline{9.03}
& \underline{0.896}
& 0.087
& 10.50/12.49
& 0.918
& 0.068
& \textcolor{green}{\ding{51}}
& \textcolor{red}{\ding{55}}
& \textcolor{red}{\ding{55}}
& \textcolor{red}{\ding{55}}
\\

StableVITON [CVPR 2024]\cite{kim2024stableviton}
& 6.8
& 7.04/9.71
& 0.872
& 0.086
& 11.57/13.67
& 0.919
& 0.066
& \textcolor{green}{\ding{51}}
& \textcolor{red}{\ding{55}}
& \textcolor{red}{\ding{55}}
& \textcolor{red}{\ding{55}}
\\

CAT-DM [CVPR 2024] \cite{zeng2024cat}
& 3.5
& 7.70/10.26
& 0.855
& 0.104
& 10.26/14.04
& 0.922
& 0.058
& \textcolor{green}{\ding{51}}
& \textcolor{red}{\ding{55}}
& \textcolor{red}{\ding{55}}
& \textcolor{red}{\ding{55}}
\\

TPD [CVPR 2024] \cite{Yang_2024_CVPR}
& 20.1
& 10.16/13.07
& 0.885
& 0.097
& 9.50/12.91
& \textbf{0.956}
& 0.048
& \textcolor{green}{\ding{51}}
& \textcolor{red}{\ding{55}}
& \textcolor{red}{\ding{55}}
& \textcolor{red}{\ding{55}}
\\

IDM-VTON [ECCV 2024] \cite{choi2024improving}
& 6.1
& 6.00/9.16
& 0.881
& 0.102
& 7.47/11.92
& 0.904
& 0.062
& \textcolor{green}{\ding{51}}
& \textcolor{red}{\ding{55}}
& \textcolor{red}{\ding{55}}
& \textcolor{red}{\ding{55}}
\\

ITA-MDT [CVPR 2025]\cite{hong2025ita}
& 3.2
& 7.66/10.66
& 0.882
& \underline{0.082}
& \underline{7.21}/11.71
& \underline{0.943}
& \textbf{0.038}
& \textcolor{green}{\ding{51}}
& \textcolor{red}{\ding{55}}
& \textcolor{red}{\ding{55}}
& \textcolor{red}{\ding{55}}
\\

CatVITON [ICLR 2025] \cite{chong2024catvton}
& 4.4
& 8.85/11.36
& 0.834
& 0.108
& 7.37/11.77
& 0.933
& 0.051
& \textcolor{green}{\ding{51}}
& \textcolor{green}{\ding{51}}
& \textcolor{red}{\ding{55}}
& \textcolor{red}{\ding{55}}
\\

PromptDresser [ICCV 2025]\cite{kim2025promptdresser}
& 3.3
& 11.15/13.06
& 0.819
& 0.151
& -
& -
& -
& \textcolor{green}{\ding{51}}
& \textcolor{red}{\ding{55}}
& \textcolor{red}{\ding{55}}
& \textcolor{red}{\ding{55}}
\\

Any2Any [ICCV 2025] \cite{guo2025any2anytryon}
& 17
& 10.04/11.09
& 0.839
& 0.090
& -
& -
& -
& \textcolor{green}{\ding{51}}
& \textcolor{red}{\ding{55}}
& \textcolor{red}{\ding{55}}
& \textcolor{red}{\ding{55}}
\\

OmniTry [NeurIPS 2025] \cite{feng2026omnitry}
& 30
& 17.08/19.92
& 0.799
& 0.202
& -
& -
& -
& \textcolor{green}{\ding{51}}
& \textcolor{green}{\ding{51}}
& \textcolor{red}{\ding{55}}
& \textcolor{red}{\ding{55}}
\\

\hline

FlexTry-On (Ours)
& \textbf{1.2}
& \textbf{5.51}/\textbf{8.97}
& 0.861
& 0.086
& \textbf{7.15}/\textbf{11.07}
& 0.927
& \underline{0.042}
& \textcolor{green}{\ding{51}}
& \textcolor{red}{\ding{55}}
& \textcolor{green}{\ding{51}}
& \textcolor{red}{\ding{55}}
\\

FlexTry-On* (Ours)
& \textbf{1.2}
& 6.34/9.22
& 0.859
& 0.093
& 7.71/\underline{11.58}
& 0.923
& 0.046
& \textcolor{green}{\ding{51}}
& \textcolor{green}{\ding{51}}
& \textcolor{green}{\ding{51}}
& \textcolor{green}{\ding{51}}
\\

\hline
\end{tabular}}
\vspace{-2mm}
\end{table*}
\subsubsection{GRAM-based Style Objective}
\label{sec:stylegram}

\begin{figure}[t]
  \centering
   \includegraphics[width=1.0\linewidth]{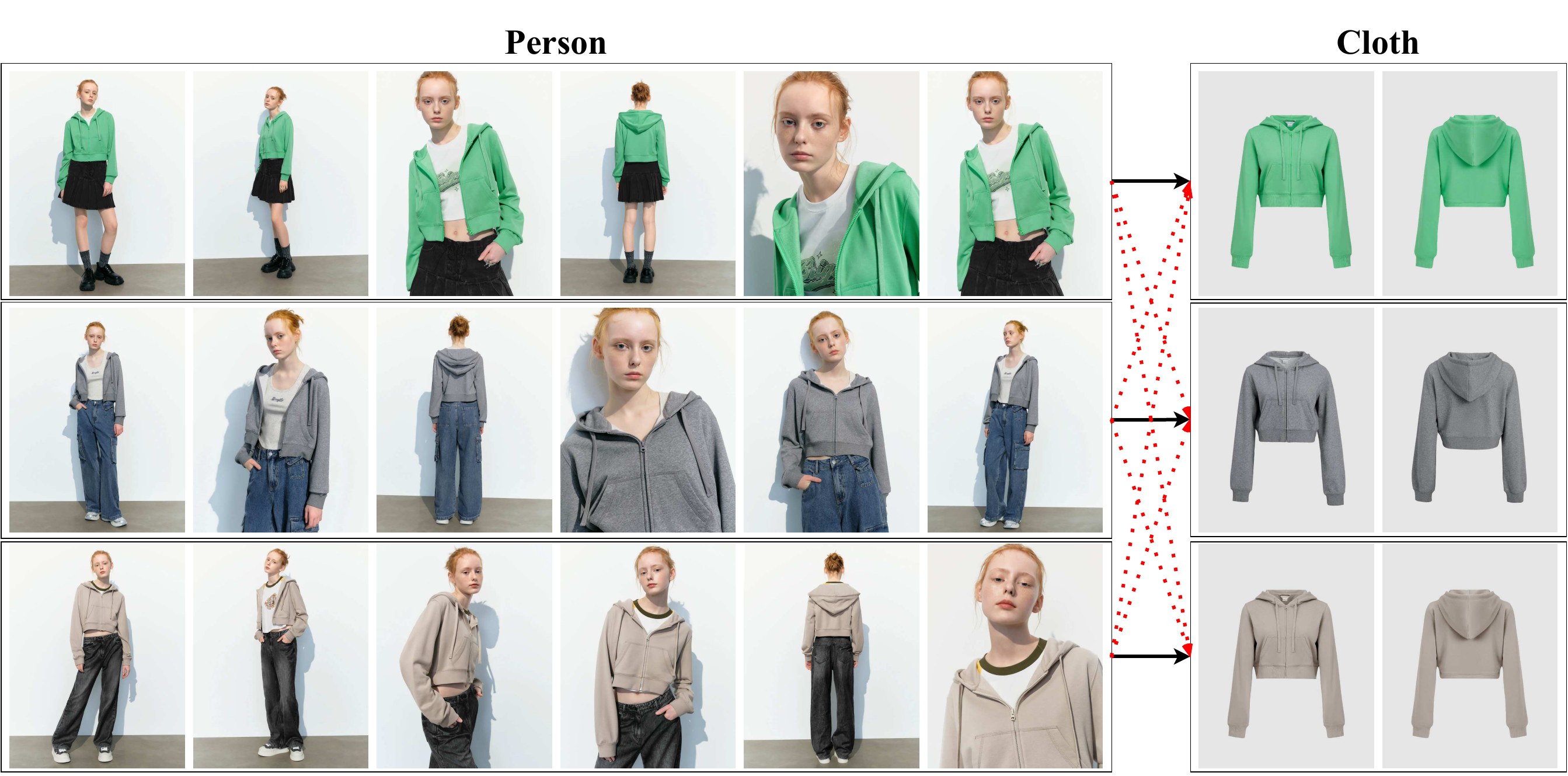}
   \vspace{-2mm}
   \captionsetup{labelfont=scriptsize,textfont=scriptsize}
   \caption{An example of mislabeled pairs in the IMAGDress dataset~\cite{shen2025imagdressing}. The black arrow indicates the correct pairing, whereas the dataset incorrectly associates each person's image with all garment images, as shown by red dashed arrows.}
   \vspace{-6mm}
   \label{fig:wronglabel}
\end{figure}

A fundamental requirement of IVTON is that the target garment \emph{must} appear in the final synthesized result. However, the diffusion denoising objective primarily supervises image reconstruction and does not explicitly optimize garment fidelity, particularly on large-scale datasets with noisy garment-person correspondences~\cite{shen2025imagdressing}. Specifically, samples may share the same garment category while differing substantially in color, texture, or fine-grained appearance. Under such conditions, the model can satisfy the denoising objective while gradually reducing its reliance on garment-specific cues, leading to degraded garment fidelity. An illustrative mis-pairing example is provided in Fig. \ref{fig:wronglabel}.

To bridge this gap, we introduce a GRAM-based Style Loss as a task-alignment constraint that explicitly encourages preservation of garment-specific appearance cues. Although GRAM-based style losses are not new~\cite{gatys2015neural}, our contribution lies in revisiting them as an optimization objective for virtual try-on rather than image stylization. Unlike pixel-wise reconstruction losses, which are incompatible with the geometric variations inherent in virtual try-on, the proposed loss operates on appearance statistics and remains invariant to spatial layout, making it particularly effective for preserving garment color and texture during synthesis. The style loss is defined as:

\begin{equation}
\mathcal{L}_{style} = \mathrm{MSE}\Big( G(z_g), \, G(z_{out} \odot M_{ic}) \Big),
\end{equation}
where $G(\cdot)$ denotes the Gram matrix, $z_g$ represents the latent garment condition, and $z_{out} \odot M_{ic}$ extracts the predicted garment region from the denoised latent.

This loss explicitly enforces the diffusion model to preserve garment-related appearance statistics throughout the synthesis process, ensuring that the generated output remains consistent with the reference garment in terms of color and texture. Unlike reconstruction-oriented objectives used in prior work \cite{ren2020deep,cui2021dressing}, the proposed GRAM-based loss does not aim to recover exact pixel values; instead, it serves as a \textbf{task-aligned constraint} that directly reflects the essential requirement of IVTON. As discussed in Sec. \ref{abl_style_loss}, this design improves garment appearance preservation under both curated and noisy training, with substantially larger gains under noisy supervision, and helps maintain garment appearance under pose and identity variations. The total loss is formulated as:
\begin{equation}
\mathcal{L}_{total} = \mathcal{L}_{denoise} + \alpha_{style} \mathcal{L}_{style},
\end{equation}
where $\alpha_{style}$ is the balance coefficient for the style loss.

\section{Experiments}
\label{sec:exp}
\begin{figure*}[ht]
  \centering
  
   \includegraphics[width=1.0\linewidth]{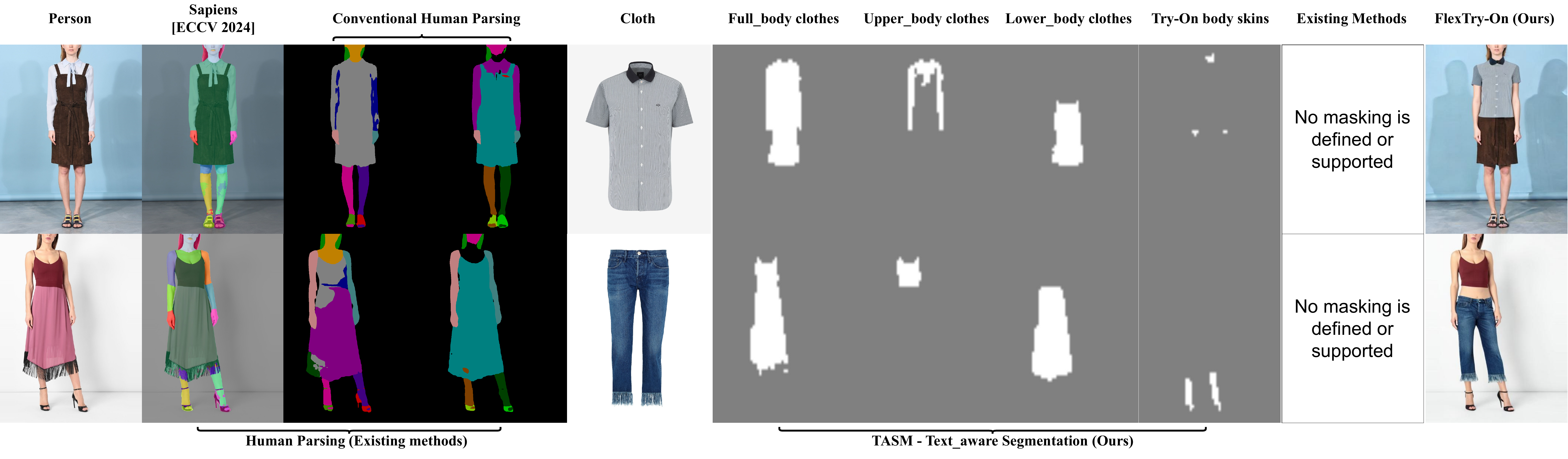}
   \vspace{-6mm}
   \captionsetup{labelfont=scriptsize,textfont=scriptsize}
   \caption{Cross-domain IVTON results on DressCode \cite{morelli2022dresscode}. Existing methods \cite{chong2024catvton,choi2024improving,hong2025ita,han2018viton,dam2024time,kim2024stableviton,kim2025promptdresser,morelli2023ladi} typically rely on category-specific human parsing \cite{li2020self} or human-centric segmentation models such as Sapiens \cite{khirodkar2024sapiens} (e.g., Voost \cite{lee2025voost}), which become ambiguous when defining editable regions in cross-domain scenarios. In contrast, our framework employs TASM for text-conditioned segmentation, enabling task-aware mask generation across different garment categories and supporting cross-domain IVTON. The TASM module operates in latent space conditioned on text; masks are interpolated for visualization purposes.}
   \label{fig:cross-domain}
   \vspace{-2mm}
\end{figure*}

\begin{figure*}[t]
  \centering
  
   \includegraphics[width=1.0\linewidth]{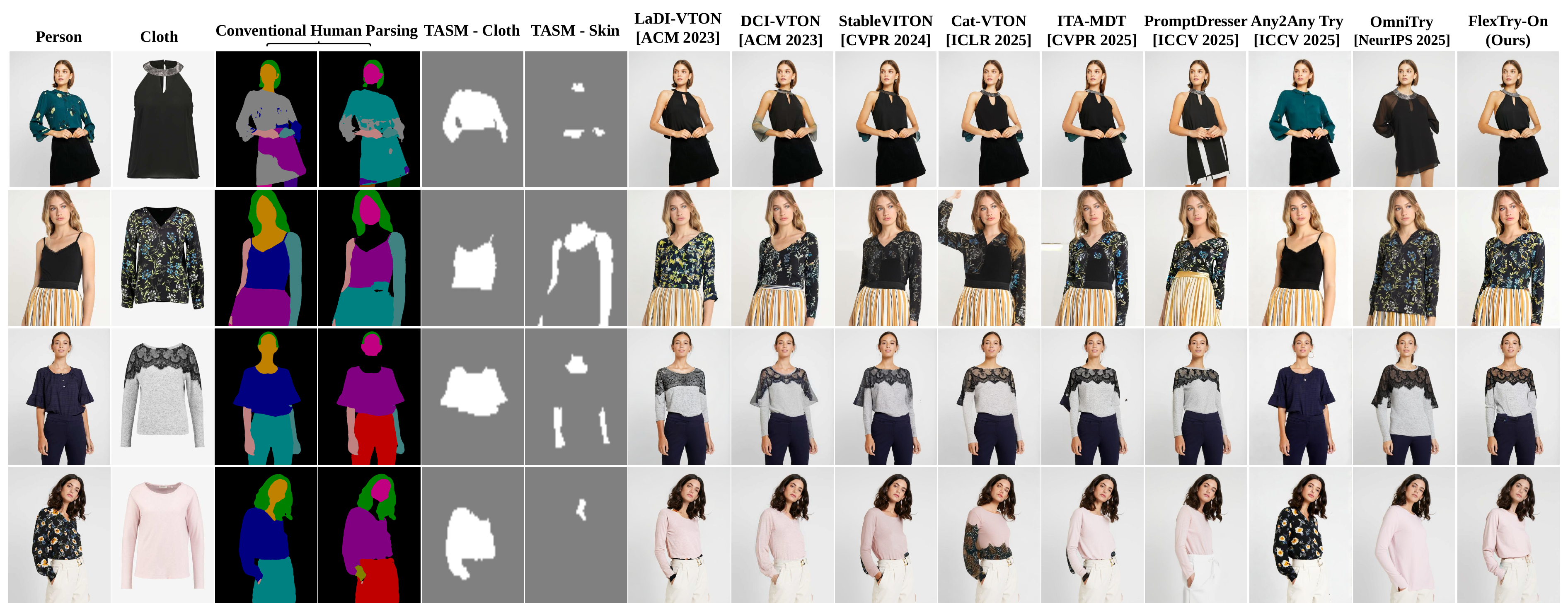}
   \captionsetup{labelfont=scriptsize,textfont=scriptsize}
   \vspace{-6mm}
   \caption{ Qualitative comparison with state-of-the-art methods on
VITON-HD~\cite{choi2021viton}. All methods are evaluated on the same test
inputs using their publicly released checkpoints when available. Conventional human parsing \cite{li2020self} may introduce inaccurate garment boundaries and incomplete editable regions, which can affect downstream virtual try-on performance. In contrast, TASM generates task-aware masks specifically designed for virtual try-on conditioning, resulting in improved garment preservation and reduced visual artifacts. Note that the TASM module operates in latent space conditioned on text; masks are interpolated for visualization purposes.
   }
   \label{fig:results}
   \vspace{-2mm}
\end{figure*}
\subsection{Experimental Setting}
\subsubsection{Implementation details}

The TASM and DDTM are trained in separate stages. Detailed configurations are provided in Appendix \ref{sec:imp_detail}


\subsubsection{Dataset and Evaluation Metrics}

For multi-scenario training, we combine
VITON-HD~\cite{choi2021viton},
DressCode~\cite{morelli2022dresscode},
DeepFashion~\cite{liu2016deepfashion},
IMAGDress~\cite{shen2025imagdressing}, and a subset of
LRVS-Fashion~\cite{lepage2023lrvsf}.
\resubmit{We denote the model trained on this full corpus as FlexTry-On*,
while FlexTry-On in Tab.~\ref{tab:fullresult} is trained only on
VITON-HD and DressCode for fair C2P comparison.}

The full corpus contains noisy or weakly aligned garment--person pairs,
motivating the task-aligned constraint in Sec.~\ref{sec:stylegram}.
Details are provided in Appendix~\ref{sec:dataset}.

Following LaDI-VTON~\cite{morelli2023ladi}, we report
FID~\cite{heusel2017gans}, SSIM~\cite{wang2004image}, and
LPIPS~\cite{zhang2018unreasonable}.
\resubmit{We additionally evaluate inference-seed sensitivity using seeds
$\{0,1,2\}$ with the same C2P checkpoint; seed-wise results are reported in
Appendix Tab.~\ref{tab:fid_seeds}(b).}

\begin{figure}[t]
  \centering
   \includegraphics[width=1.0\linewidth]{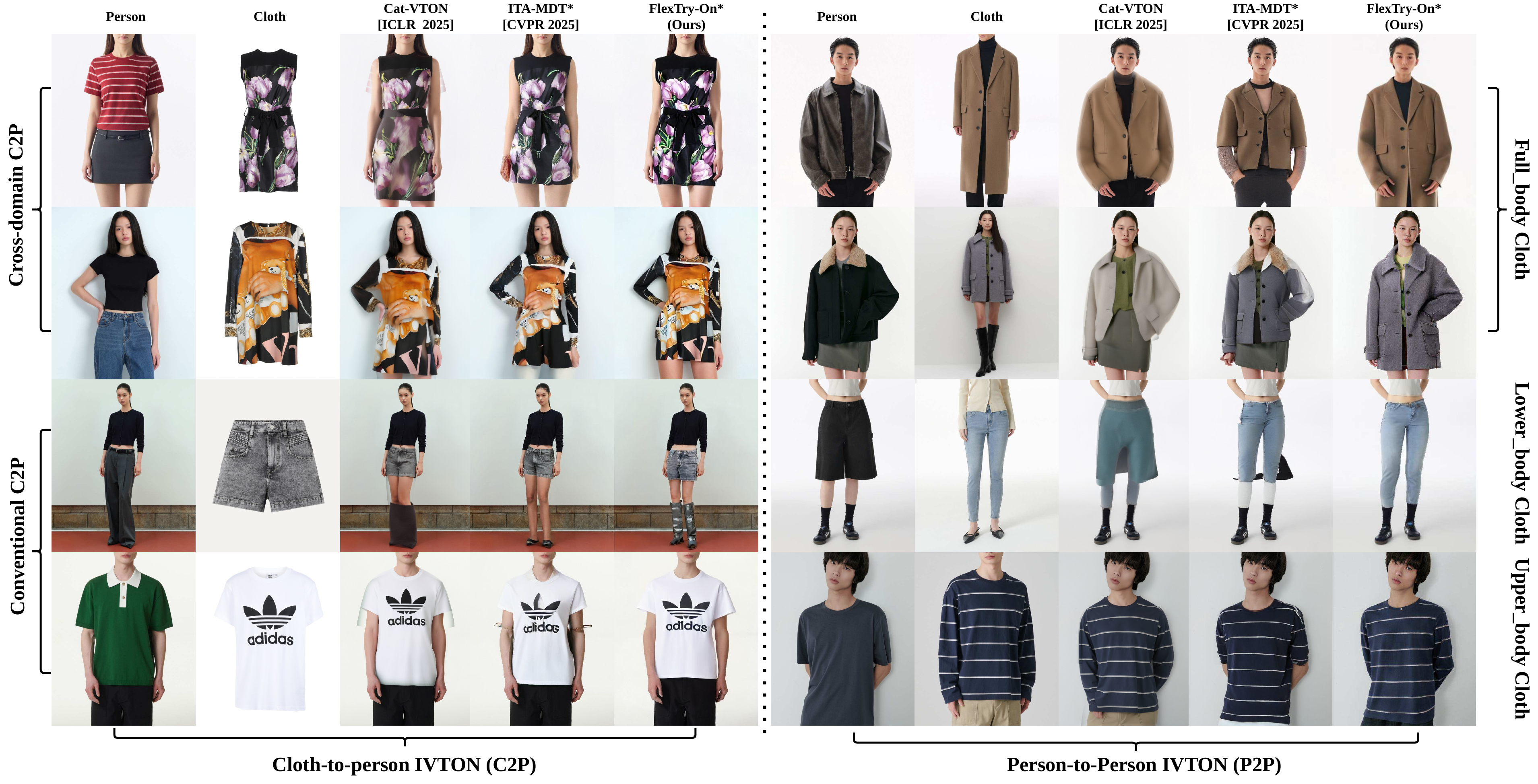}
   \vspace{-6mm}
   \captionsetup{labelfont=scriptsize,textfont=scriptsize}
   \caption{Qualitative results in C2P and P2P IVTON setting. The person and clothes images are collected from the Internet. FlexTry-On* is our proposed method trained on merged uncurated datasets. ITA-MDT* denotes the fine-tuned ITA-MDT with segmented clothing, trained on our dataset (the full, uncurated dataset).
   }
   \label{fig:wild3}
   \vspace{-8mm}
\end{figure}

\begin{table*}[ht]
\centering
\captionsetup{labelfont=scriptsize,textfont=scriptsize}
\caption{
Ablation study quantitative results on VITON-HD
\cite{choi2021viton} and DressCode \cite{morelli2022dresscode} Upper in C2P.
DDTM denotes the DDTM trained only on VITON-HD and DressCode in C2P,
whereas DDTM* denotes training on the full uncurated combined corpus.
Best/second-best results are \textbf{bold}/\underline{underlined}.
Times are measured for one batch-size-1 try-on. \resubmit{For that image, each
segmenter evaluates all 10 segmentation-class conditions listed in
Appendix~\ref{sec:tasm_training_config}. TASM batches them in 0.1~s, whereas
G-SAM runs GroundingDINO followed by SAM sequentially for each condition
(1.9~s per class; 19.0~s total). Denoising takes 1.1~s per try-on.}
}
\vspace{-1mm}
\label{tab:abl_main}

\renewcommand{\arraystretch}{1.12}
\resizebox{\textwidth}{!}{
\begin{tabular}{l|c|cc|ccc|ccc}
\hline

\multicolumn{1}{c|}{} &
\multicolumn{1}{c|}{Seg Param} &
\multicolumn{2}{c|}{\resubmit{Time / try-on}} &
\multicolumn{3}{c|}{VITON-HD} &
\multicolumn{3}{c}{DressCode - Upper} \\

\multicolumn{1}{c|}{Method} &
(M Param) $\downarrow$ &
\resubmit{Seg. (s) $\downarrow$} &
\resubmit{Denoising (s) $\downarrow$} &
FID (pair/unp.) $\downarrow$ &
SSIM $\uparrow$ &
LPIPS $\downarrow$ &
FID (pair/unp.) $\downarrow$ &
SSIM $\uparrow$ &
LPIPS $\downarrow$ \\

\hline

DDTM (Baseline)
& 65
& 0.1
& 1.1
& 6.21/9.64
& 0.857
& 0.091
& 8.89/12.84
& 0.911
& 0.061
\\

Baseline + TASM
& 130
& 0.1
& 1.1
& \underline{5.83}/\underline{9.02}
& \textbf{0.861}
& \textbf{0.085}
& \textbf{7.12}/\underline{11.11}
& \underline{0.925}
& \textbf{0.042}
\\

Baseline + TASM + $\mathcal{L}_{style}$
& 130
& 0.1
& 1.1
& \textbf{5.51}/\textbf{8.97}
& \textbf{0.861}
& \underline{0.086}
& \underline{7.15}/\textbf{11.07}
& \textbf{0.927}
& \textbf{0.042}
\\

\hline

Baseline + G-SAM
& 136
& 19.0
& 1.1
& 7.15/10.27
& 0.841
& 0.107
& -
& -
& -
\\

\hline

DDTM* + TASM
& 130
& 0.1
& 1.1
& 6.80/9.42
& 0.857
& 0.093
& 8.22/11.92
& 0.923
& 0.048
\\

DDTM* + TASM + $\mathcal{L}_{style}$
& 130
& 0.1
& 1.1
& 6.34/9.22
& \underline{0.859}
& 0.093
& 7.71/11.58
& 0.923
& \underline{0.046}
\\

\hline
\end{tabular}
}

\vspace{-4mm}
\end{table*}

\subsection{Quantitative and Qualitative Evaluation}

\subsubsection{Baseline Comparison}

Tab.~\ref{tab:fullresult} compares FlexTry-On with prior diffusion-based
IVTON methods. Publicly released checkpoints are used whenever available.
\resubmit{For runtime evaluation, FlexTry-On uses 30 denoising steps, whereas
each baseline follows its default inference configuration. All directly
reproducible methods are measured with batch size 1 on a single RTX 4090
(24GB VRAM); exceptions and hardware normalization are detailed in
Appendix~\ref{sec:imp_detail}.} All evaluations are conducted at
$512\times512$ resolution. For methods without official implementations,
we report paper-reported results. For OmniVTON~\cite{yang2025omnivton},
we attempted to reproduce the official implementation but encountered
unresolved issues and therefore retain its published quantitative results.

FlexTry-On achieves the best paired/unpaired FID on both VITON-HD and
DressCode-Upper among the evaluated methods while maintaining fast inference.
Its efficiency comes from a single Reference-Net pass for garment feature
extraction, followed by iterative denoising in the Main Net, avoiding repeated
garment-side computation across denoising steps. Qualitative results in
Fig.~\ref{fig:results} further show more consistent garment texture and
structure preservation, with fewer arm-region artifacts and clothing
inconsistencies.


\subsubsection{Full Dataset Training}

To support multiple IVTON scenarios within a single model, we train
FlexTry-On* on the full combined C2P--P2P corpus. As shown in
Tab.~\ref{tab:fullresult}, FlexTry-On* supports C2P, P2P, and cross-domain
C2P/P2P, providing broader scenario coverage than the compared methods.
We evaluate its standard C2P performance on the VITON-HD
~\cite{choi2021viton} and DressCode~\cite{morelli2022dresscode} Upper
test sets.

\resubmit{Compared with FlexTry-On trained only on VITON-HD and DressCode,
FlexTry-On* shows a modest reduction in standard C2P performance
(5.51/8.97 vs.\ 6.34/9.22 paired/unpaired FID on VITON-HD)}, while enabling
P2P and cross-domain scenarios within the same model. This illustrates the
trade-off between optimizing for a single benchmark setting and supporting
broader scenarios.

\resubmit{As shown in Fig.~\ref{fig:wild3}, FlexTry-On* supports P2P in both
same-domain and zero-shot cross-domain settings.} We further compare it with
CatVTON and ITA-MDT*, a retrained variant of ITA-MDT. For cross-domain
dress transfer, these methods require manually defined full-garment masks
to specify the editable region because their original masking strategies
are designed for category-consistent try-on. Even with these manually
adapted masks, preserving garment structure and appearance across categories
remains challenging. Meanwhile, TASM automatically constructs task-aware
editable regions, allowing FlexTry-On* to produce more coherent results
with better garment texture and structure preservation. Additional
cross-domain examples, including dress-to-lower, dress-to-upper, and
upper-lower-to-dress transfers in C2P/P2P, are provided
in Appendix~\ref{sec:Crossdomain_discussion}.

\begin{figure}[t]
  \centering
   \includegraphics[width=1.0\linewidth]{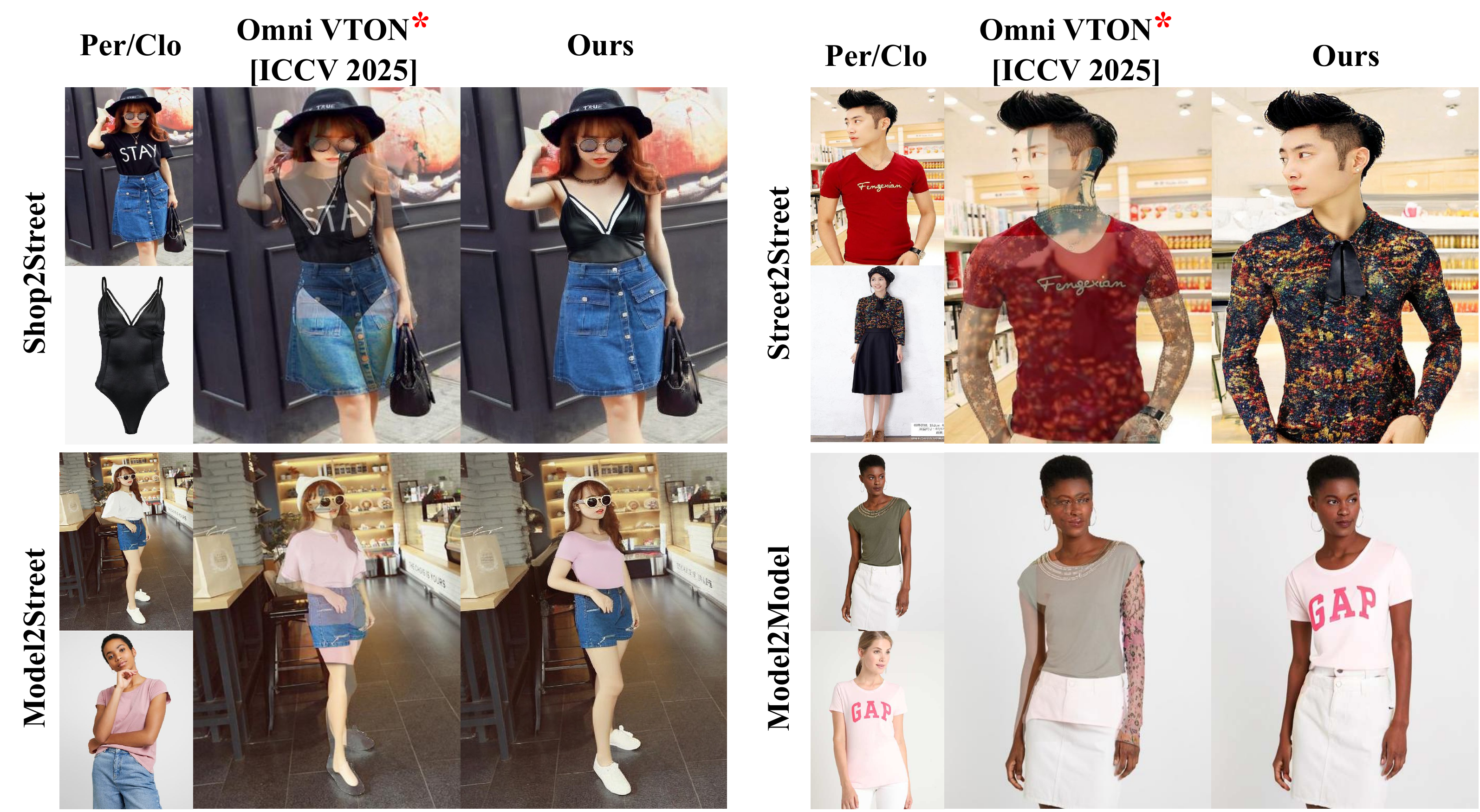}
   \vspace{-6mm}
   \captionsetup{labelfont=scriptsize,textfont=scriptsize}
   \caption{Qualitative StreetTryOn results~\cite{cui2025street}. \textcolor{red}{*} \resubmit{marks our unsuccessful reproduction.}}
   \label{fig:street}
   \vspace{-2mm}
\end{figure}
\begin{table}[t]
\centering
\captionsetup{labelfont=scriptsize,textfont=scriptsize}
\caption{Results on Street Try-On. $\dagger$: OmniVTON \cite{yang2025omnivton} reported the result
}
\vspace{-2mm}
\label{tab:street}

\renewcommand{\arraystretch}{1.08}
\resizebox{\linewidth}{!}{
\begin{tabular}{l|cccc} 
\hline
\multicolumn{1}{c|}{}  &   \multicolumn{1}{c}{Shop2Street}  & \multicolumn{1}{c}{Model2Model} & \multicolumn{1}{c}{Model2Street} & \multicolumn{1}{c}{Street2Street} \\
\multicolumn{1}{c|}{Method}   & \multicolumn{1}{c}{FID↓} & \multicolumn{1}{c}{FID↓} & \multicolumn{1}{c}{FID↓} & \multicolumn{1}{c}{FID↓} \\ 
\hline
CAT-DM $\dagger$[CVPR 2024] \cite{zeng2024cat}                 & 37.48                            & -                            & -                      & -     \\
IDM-VTON $\dagger$[ECCV 2024] \cite{choi2024improving}                 & 42.28                            & -                            & -                      & -    \\
OmniVTON $\dagger$[ICCV 2025] \cite{yang2025omnivton}                & 33.92                            & \textbf{8.98}                            & 33.45                      & \textbf{23.47}     \\                                           
 
FlexTry-On (Ours)  & \textbf{32.51} & 10.06       & \textbf{33.07}                         & 32.6                     \\

\hline
\end{tabular}}
\vspace{-4mm}
\end{table}


\subsubsection{In-the-wild Evaluation}

On Street Try-On~\cite{cui2025street}, both FlexTry-On and OmniVTON are
evaluated zero-shot, without training on StreetTryOn. As shown in
Tab.~\ref{tab:street}, FlexTry-On performs better on Shop2Street and
Model2Street, whereas the paper-reported OmniVTON results are better on
Model2Model and Street2Street. Street2Street remains challenging
due to uncontrolled pose, illumination, background, and occlusion, especially
when the source garment is small or partially visible (Appendix~H).

\resubmit{For OmniVTON~\cite{yang2025omnivton}, we retain the published
StreetTryOn scores for comparison. However, our evaluation using the official
repository did not reproduce the reported performance and produced degraded
outputs (Fig.~\ref{fig:street}). This discrepancy suggests that the results
may depend on implementation details or evaluation conditions not
captured by the released setup. Thus, these results are paper-reported rather
than independently verified, and the reported inference times are not directly
comparable across environments.}

\begin{figure*}[t]
  \centering
   \includegraphics[width=1.0\linewidth]{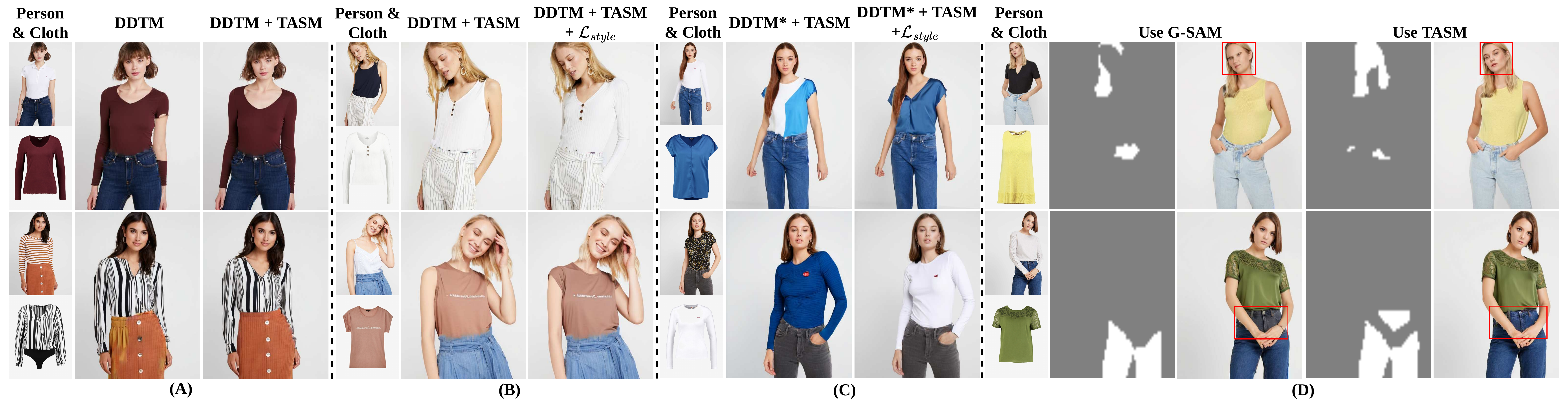}
   \vspace{-6mm}
   \captionsetup{labelfont=scriptsize,textfont=scriptsize}
   \caption{Qualitative ablation on VITON-HD~\cite{choi2021viton} under curated and full noisy-data training. DDTM and DDTM* denote the two settings, respectively; ``Use G-SAM'' replaces TASM with Grounded-SAM under matched-text conditions.
   }
   \label{fig:abl}
   \vspace{-6mm}
\end{figure*}

\begin{figure}[t]
  \centering
   \includegraphics[width=1.0\linewidth]{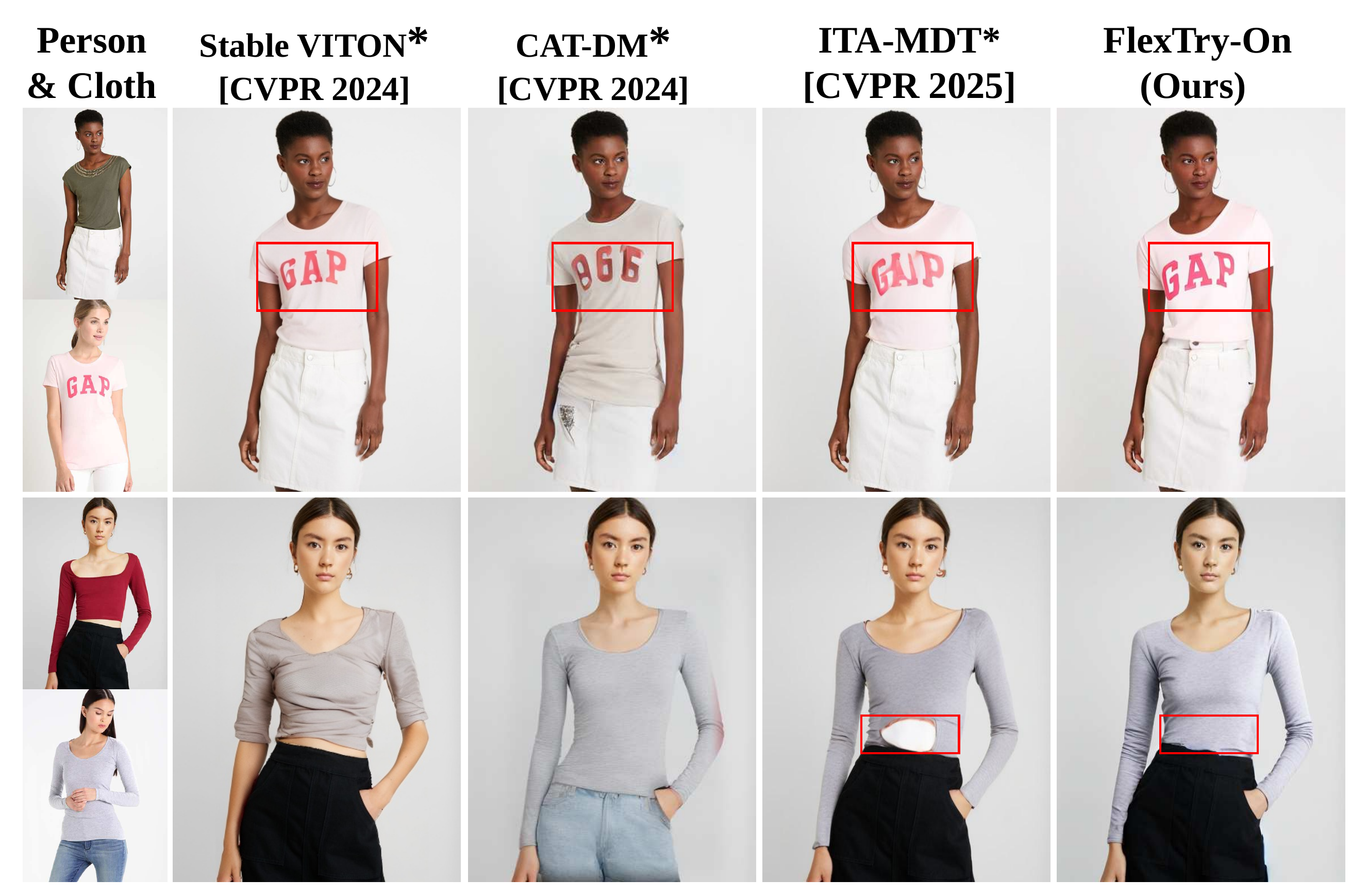}
   \vspace{-6mm}
   \captionsetup{labelfont=scriptsize,textfont=scriptsize}
   \caption{Qualitative result on VITON-HD, P2P setting. Model*: retrained model.}
   \label{fig:P2P}
   \vspace{-2mm}
\end{figure}

\subsection{Ablation Studies}

\subsubsection{Effectiveness of TASM}

We compare TASM with conventional human parsing and Grounded-SAM (G-SAM)~\cite{liu2023grounding,ren2024grounded}. Because reliable ground-truth masks are unavailable, particularly for cross-domain cases, and mask accuracy does not directly measure conditioning quality, we evaluate downstream try-on performance rather than Boundary IoU~\cite{cheng2021boundary}.

Replacing TASM degrades quantitative and visual quality (Tab.~\ref{tab:abl_main}, Fig.~\ref{fig:abl}(A)(D)), producing incomplete garment preservation and retained body regions. Although trained from G-SAM pseudo-labels, TASM provides better task-aware conditioning than G-SAM. Its latent-space design also supports category-aware cross-domain masking without additional supervision; pixel-space alternatives are discussed in Sec.~\ref{sec:fastSAM}.

\subsubsection{Effectiveness of the Style-GRAM Loss}
\label{abl_style_loss}
\begin{table}[t]
\centering

\captionsetup{labelfont=scriptsize,textfont=scriptsize}

\caption{\resubmit{
P2P and C2P results on VITON-HD.
$^*$ denotes retraining on our full C2P--P2P corpus.
Best and second-best results are \textbf{bold} and \underline{underlined}.
}}

\vspace{-2mm}

\label{tab:fullresult-abl}

\renewcommand{\arraystretch}{1.10}

\resizebox{\linewidth}{!}{
\begin{tabular}{l|c|ccc|ccc}
\hline

\multicolumn{1}{c|}{} &
\multicolumn{1}{l|}{Inf. Time} &
\multicolumn{3}{c|}{VITON-HD \resubmit{P2P}} &
\multicolumn{3}{c}{\resubmit{VITON-HD C2P}} \\

\multicolumn{1}{c|}{Method} &
(s) $\downarrow$ &
\multicolumn{1}{l}{FID (pair/unp.) $\downarrow$} &
\multicolumn{1}{l}{SSIM $\uparrow$} &
\multicolumn{1}{l|}{LPIPS $\downarrow$} &
\multicolumn{1}{l}{\resubmit{FID (pair/unp.) $\downarrow$}} &
\multicolumn{1}{l}{\resubmit{SSIM $\uparrow$}} &
\multicolumn{1}{l}{\resubmit{LPIPS $\downarrow$}} \\

\hline

Stable VITON* [CVPR 2024] \cite{kim2024stableviton}
& 6.8
& \underline{7.01}/\underline{10.35}
& 0.876
& 0.098
& \underline{6.93}/\underline{9.53}
& 0.856
& \textbf{0.091}
\\

CAT-DM* [CVPR 2024] \cite{zeng2024cat}
& 3.5
& 10.71/12.44
& 0.832
& 0.153
& 8.72/11.32
& 0.843
& 0.124
\\

ITA-MDT* [CVPR 2025] \cite{hong2025ita}
& \underline{3.2}
& 8.02/11.91
& \underline{0.881}
& \underline{0.078}
& 7.99/11.34
& \textbf{0.876}
& 0.094
\\

\textbf{FlexTry-On* (Ours)}
& \textbf{1.2}
& \textbf{4.01}/\textbf{10.06}
& \textbf{0.905}
& \textbf{0.044}
& \textbf{6.34}/\textbf{9.22}
& \underline{0.859}
& \underline{0.093}
\\

\hline
\end{tabular}
}

\vspace{-4mm}
\end{table}

Tab.~\ref{tab:abl_main} isolates $\mathcal{L}_{style}$ under matched training
settings. On curated VITON-HD and DressCode, it maintains comparable image
quality while slightly improving unpaired FID. Notably, on DressCode-Upper,
paired FID changes only from 7.12 to 7.15, whereas unpaired FID improves
from 11.11 to 11.07.

The benefit becomes clearer under noisy full-data training.
On VITON-HD, $\mathcal{L}_{style}$ improves paired/unpaired FID from
6.80/9.42 to 6.34/9.22; on DressCode-Upper, it improves from
8.22/11.92 to 7.71/11.58, while maintaining comparable SSIM/LPIPS.
Together with the qualitative improvements in Fig.~\ref{fig:abl}(B,C),
these results support $\mathcal{L}_{style}$ as a garment-preservation
constraint, particularly under noisy supervision.

\subsubsection{Effect of Extended Training Data on Prior Methods}

\resubmit{To separate model design from data scale, we retrain StableVITON,
CAT-DM, and ITA-MDT on the same C2P--P2P corpus, adapting their released
checkpoints with segmented person garments while following the original
protocols. Methods requiring additional conditioning are excluded.
Tab.~\ref{tab:fullresult-abl} and Fig.~\ref{fig:P2P} evaluate the resulting
checkpoints in both settings.}

With matched training data, FlexTry-On achieves the best paired/unpaired FID
in both C2P and P2P.Qualitative results in Fig.~\ref{fig:P2P}
further support this result, showing more consistent garment appearance and
fewer distortions under pose and identity changes. Taken together, these
controlled comparisons indicate that the observed gains cannot be explained
by the broader training corpus alone, but also reflect the effectiveness of
the proposed unified conditioning and generation framework.

\section{\resubmit{Limitations}}
\label{sec:limitations}

\resubmit{FlexTry-On remains challenged by severe occlusion, unusual or oversized
garments, and Street-to-Street cases involving uncontrolled pose,
illumination, background, and occlusion. The current pipeline assumes a
single selected visible person and does not automatically handle crowded or
multi-person scenes. The combined training corpus may also inherit demographic
and social-media biases.}

\resubmit{In addition, FlexTry-On does not consistently outperform all baselines on
SSIM/LPIPS, despite achieving stronger FID in several settings. The OmniVTON
comparison relies on paper-reported quantitative results because our local
reproduction was unsuccessful. Additional discussion and failure analysis are
provided in Appendices~\ref{sec:suppl_discussion}
and~\ref{sec:suppl_limitations}. }

\resubmit{The full training corpus also contains weakly aligned garment--person pairs;
$\mathcal{L}_{style}$ mitigates their effect on garment preservation but does
not eliminate supervision noise.}

\resubmit{In practice, FlexTry-On targets single-person try-on applications where a
target person and garment are specified before inference; automatic person
selection in crowded scenes is outside the current scope.}
\section{Conclusion}
\label{sec:end}
We present FlexTry-On, a unified framework for C2P, P2P, cross-domain,
and in-the-wild virtual try-on. TASM provides task-aware editable regions,
DDTM enables efficient synthesis with a single garment-reference pass, and
$\mathcal{L}_{style}$ improves garment preservation under noisy supervision.
FlexTry-On achieves competitive quality with 1.2-s inference, while matched
retraining and qualitative results support its effectiveness across diverse
try-on settings.

\section*{Acknowledgements}

This work was supported by the National Research Foundation of Korea (NRF) grant funded by the Korean government (MSIT) (RS-2025-00573160) and by the “Advanced GPU Utilization Support Program” funded by the Government of the Republic of Korea (Ministry of Science and ICT).
{
    \small
    \bibliographystyle{ieeenat_fullname}
    \bibliography{main}
}
\clearpage
\appendix

\renewcommand{\thesection}{\Alph{section}}
\renewcommand{\thefigure}{\Alph{figure}}
\renewcommand{\thetable}{\Alph{table}}

\setcounter{page}{1}
\setcounter{section}{0}
\setcounter{figure}{0}
\setcounter{table}{0}

\begin{center}
{\LARGE\bfseries Supplementary Material}\\[1em]
\end{center}
\begin{table}[ht]
\centering
\vspace{-3mm}
\captionsetup{labelfont=scriptsize,textfont=scriptsize}
\caption{Quantitative results of diffusion-based virtual try-on methods on VITON-HD and DressCode upper-body test sets. We report both paired and unpaired FID. TASM-MaskOnly denotes the TASM with output that is the mask only. DDTM refers to the training of the Dual-DiT Tryon Module with only VITON-HD \cite{choi2021viton} and Dresscode \cite{morelli2022dresscode} in the C2P tasks. The better performance for each metric is \textbf{bold}.}
\vspace{-3mm}
\label{tab:tasm-abl}
\resizebox{\linewidth}{!}{
\begin{tabular}{l|cccccc} 
\hline
\multicolumn{1}{l|}{}     & \multicolumn{3}{c}{VITON-HD}                                                                     & \multicolumn{3}{c}{Dress Code - Upper}                                                           \\
\multicolumn{1}{c|}{Method}                     & \multicolumn{1}{l}{FID (pair/unp.) ↓} & \multicolumn{1}{l}{SSIM ↑} & \multicolumn{1}{l}{LPIPS ↓} & \multicolumn{1}{l}{FID (pair/unp.)↓} & \multicolumn{1}{l}{SSIM ↑} & \multicolumn{1}{l}{LPIPS ↓}  \\ 
\hline
DDTM + TASM-MaskOnly +  $\mathcal{L}_{style}$ & 5.98/9.08                                  & 0.860                         & 0.086                          & 7.20/11.08                                & 0.924                          & 0.043                            \\
DDTM + TASM +  $\mathcal{L}_{style}$ & \textbf{5.51}/\textbf{8.97}       & \textbf{0.861}                          & 0.086                           & \textbf{7.15}/\textbf{11.07}                                   & \textbf{0.927}                          & \textbf{0.042}                            \\
\hline
\end{tabular}}
\vspace{-4mm}
\end{table}

\section{Additional Ablation Study}
\subsection{Empirical Justification of the TASM Design}
\label{sec:TASM_design_choice}
This ablation investigates a key design choice of TASM. Although TASM predicts both latent appearance representations and segmentation masks, only the masks are used during try-on inference. The latent representations serve solely as an auxiliary supervision signal during training, raising the question about this branch's necessary. 

\begin{figure}[t]
  \centering
   \includegraphics[width=1.0\linewidth]{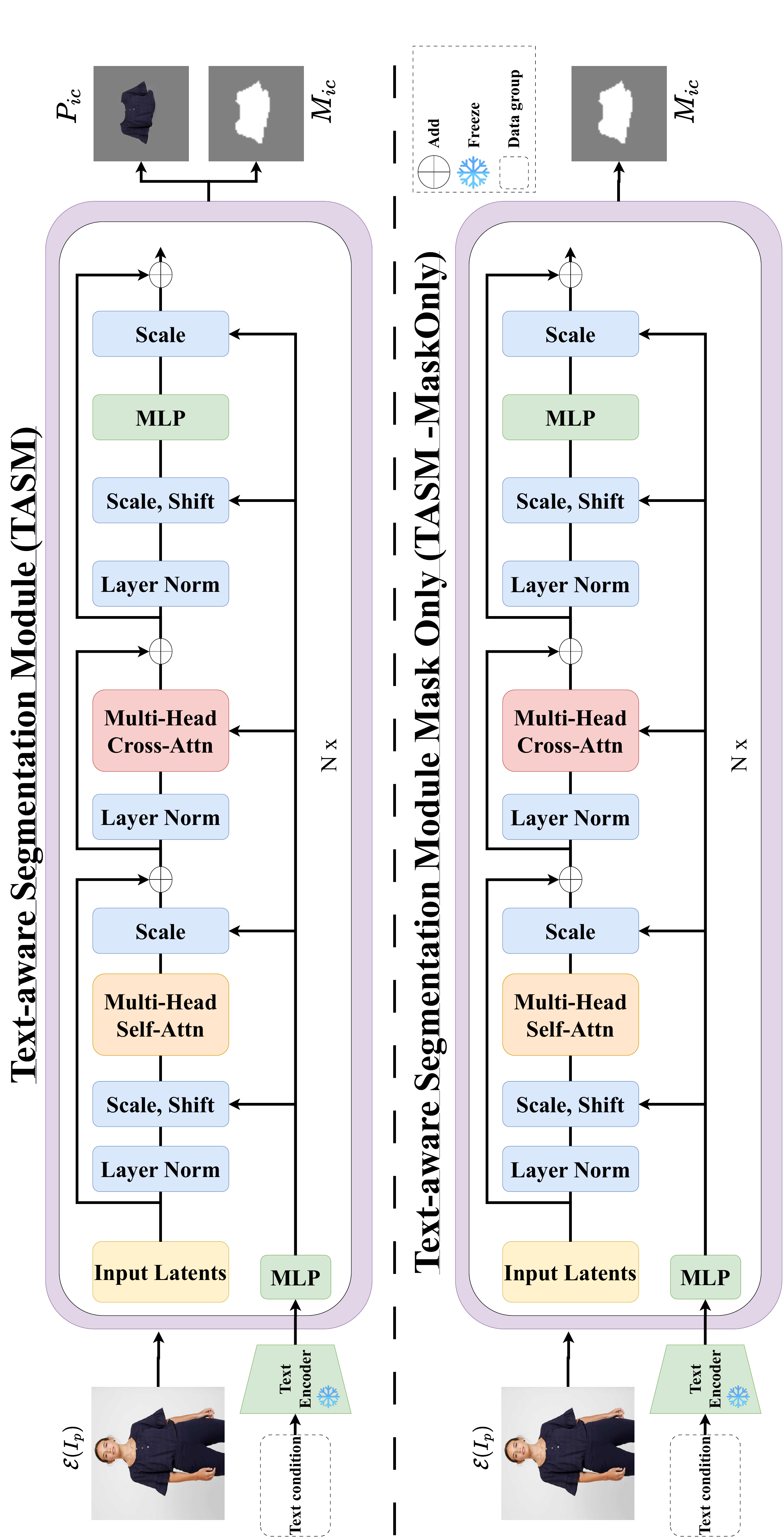}
    \captionsetup{labelfont=scriptsize,textfont=scriptsize}
   \caption{Overview of the Text-aware segmentation Module for FlexTry-On Framework and its variant. The TASM module receives a latent image with text conditions embedded via the CLIP-Text Encoder from Stable Diffusion~2.0~\cite{rombach2022high}. The module outputs both the latent representations of the target RGB regions and their corresponding masks. In TASM, we illustrate the latent representation of the RGB input cloth $P_{ic}$ and its associated mask $M_{ic}$. While in TASM-MaskOnly, the output is only the mask $M_{ic}$.}
   \label{fig:textseg-apd}
   \vspace{-6mm}
\end{figure}

To answer this question, we construct a simplified variant, denoted as \textit{TASM-MaskOnly}, which predicts segmentation masks only while retaining the same architecture and training objective. An overview of both designs is shown in Fig.~\ref{fig:textseg-apd}. As described in the main paper, TASM injects text conditions through multi-head cross-attention and scale-and-shift modulation. Both variants are integrated into the same Dual-DiT framework with the Style-GRAM loss, trained on VITON-HD and DressCode in the C2P setting and evaluated on their respective Upper test sets.

We hypothesize that latent prediction acts as an auxiliary representation-learning task. While mask supervision mainly focuses on region boundaries, reconstructing latent garment representations requires the model to capture garment semantics, appearance, and spatial extent simultaneously. Consequently, the learned features become more informative for segmentation, potentially leading to more accurate mask prediction even though the latent outputs are discarded during inference.

\begin{figure}[t]
    \centering
   \includegraphics[width=1.0\linewidth]{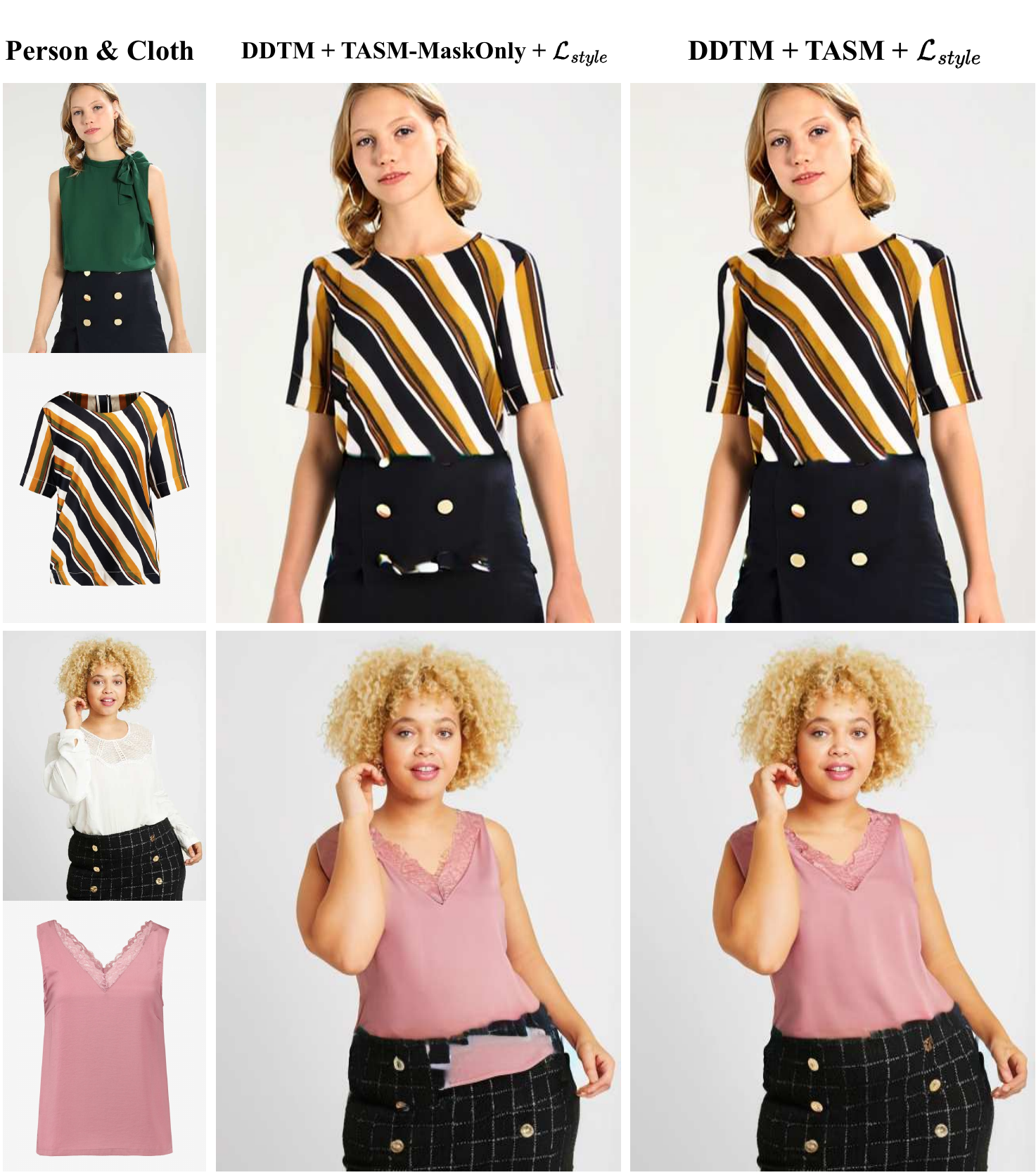}
   \vspace{-2mm}
   \captionsetup{labelfont=scriptsize,textfont=scriptsize}
   \caption{Ablation study results on VITON-HD \cite{choi2021viton} to evaluate the effectiveness of the TASM module architecture design in the clean datasets. The DDTM denotes the Dual-DiT Tryon Module, which is trained only on the clean dataset (VITON-HD \cite{choi2021viton} and DressCode \cite{morelli2022dresscode}).}
   \label{fig:tasm-maskonly}
   \vspace{-6mm}
\end{figure}

Quantitative results in Tab.~\ref{tab:tasm-abl} show that TASM consistently outperforms TASM-MaskOnly across nearly all metrics. Qualitative comparisons in Fig.~\ref{fig:tasm-maskonly} further reveal that TASM-MaskOnly frequently fails to accurately localize garment regions, particularly for lower-body garments, leading to degraded try-on results. These findings support our hypothesis that auxiliary latent prediction improves region understanding and contributes to higher-quality garment masks, despite not being used during inference.

\subsection{Further discussion about TASM}
\label{sec:fastSAM}
\textit{"Why not light-weight pixel segmentation?"} FastSAM, a lightweight SAM, is designed by sacrificing segmentation accuracy for speed relative to SAM. Our empirical advantage over the SAM-based approach further supports TASM's effectiveness. These results motivate the latent-space formulation of TASM. By focusing on semantically consistent garment regions rather than exact contours, TASM improves robustness, computational efficiency, and compatibility with diffusion-based synthesis, while enabling more flexible scenarios such as cross-domain try-on.

\section{DressCode Results}
\label{sec:quantitavedresscode}

\begin{table*}[ht]
\centering
\captionsetup{font=scriptsize,skip=2pt}
\caption{\resubmit{Additional quantitative results. (a) DressCode lower-body and dresses results; inference time denotes method-specific inference time for one image. FlexTry-On is trained on VITON-HD \cite{choi2021viton} and DressCode \cite{morelli2022dresscode} for C2P. Best and second-best results are \textbf{bold} and \underline{underlined}. (b) Seed-wise VITON-HD FID using the same FlexTry-On checkpoint; the last row reports mean $\pm$ standard deviation.}}
\label{tab:fullresult-dresscode}
\label{tab:fid_seeds}
\begin{minipage}[t]{0.72\textwidth}
\centering
{\scriptsize\textbf{(a) DressCode lower-body and dresses results}}\\[2pt]
\setlength{\tabcolsep}{4pt}
\renewcommand{\arraystretch}{1.05}
\resizebox{\linewidth}{!}{%
\begin{tabular}{l c ccc ccc}
\toprule
& & \multicolumn{3}{c}{DressCode -- Lower} & \multicolumn{3}{c}{DressCode -- Dresses} \\
\cmidrule(lr){3-5}\cmidrule(lr){6-8}
Method & Time (s) $\downarrow$ & FID (pair/unp.) $\downarrow$ & SSIM $\uparrow$ & LPIPS $\downarrow$ & FID (pair/unp.) $\downarrow$ & SSIM $\uparrow$ & LPIPS $\downarrow$ \\
\midrule
LaDI-VTON [ACM 2023] \cite{morelli2023ladi} & \underline{2.1} & 10.97/16.31 & 0.903 & 0.061 & 12.75/15.80 & 0.852 & 0.108 \\
ITA-MDT [CVPR 2025] \cite{hong2025ita} & 3.2 & \underline{8.22}/14.03 & \textbf{0.921} & \underline{0.054} & 7.79/11.51 & \underline{0.878} & 0.081 \\
CatVTON [ICLR 2025] \cite{chong2024catvton} & 4.4 & 8.26/\underline{13.41} & \underline{0.920} & \textbf{0.037} & \underline{7.66}/\textbf{10.80} & 0.862 & \textbf{0.055} \\
\midrule
FlexTry-On (Ours) & \textbf{1.2} & \textbf{7.63/13.31} & 0.906 & 0.056 & \textbf{7.05}/\underline{10.89} & \textbf{0.880} & \underline{0.075} \\
\bottomrule
\end{tabular}}
\end{minipage}
\hfill
\begin{minipage}[t]{0.25\textwidth}
\centering
\scriptsize
\textbf{\resubmit{(b) Seed-wise VITON-HD FID}}\\[2pt]
\renewcommand{\arraystretch}{1.0}
\resizebox{\linewidth}{!}{%
\begin{tabular}{c cc}
\toprule
Seed & Paired FID $\downarrow$ & Unpaired FID $\downarrow$ \\
\midrule
0 & 5.51 & 8.97 \\
1 & 5.54 & 8.99 \\
2 & 5.50 & 8.98 \\
\midrule
\textbf{Mean $\pm$ std.} & $\mathbf{5.52 \pm 0.02}$ & $\mathbf{8.98 \pm 0.01}$ \\
\bottomrule
\end{tabular}}
\end{minipage}
\vspace{-4mm}
\end{table*}

\begin{figure*}[t]
  \centering
   \includegraphics[width=0.9\linewidth]{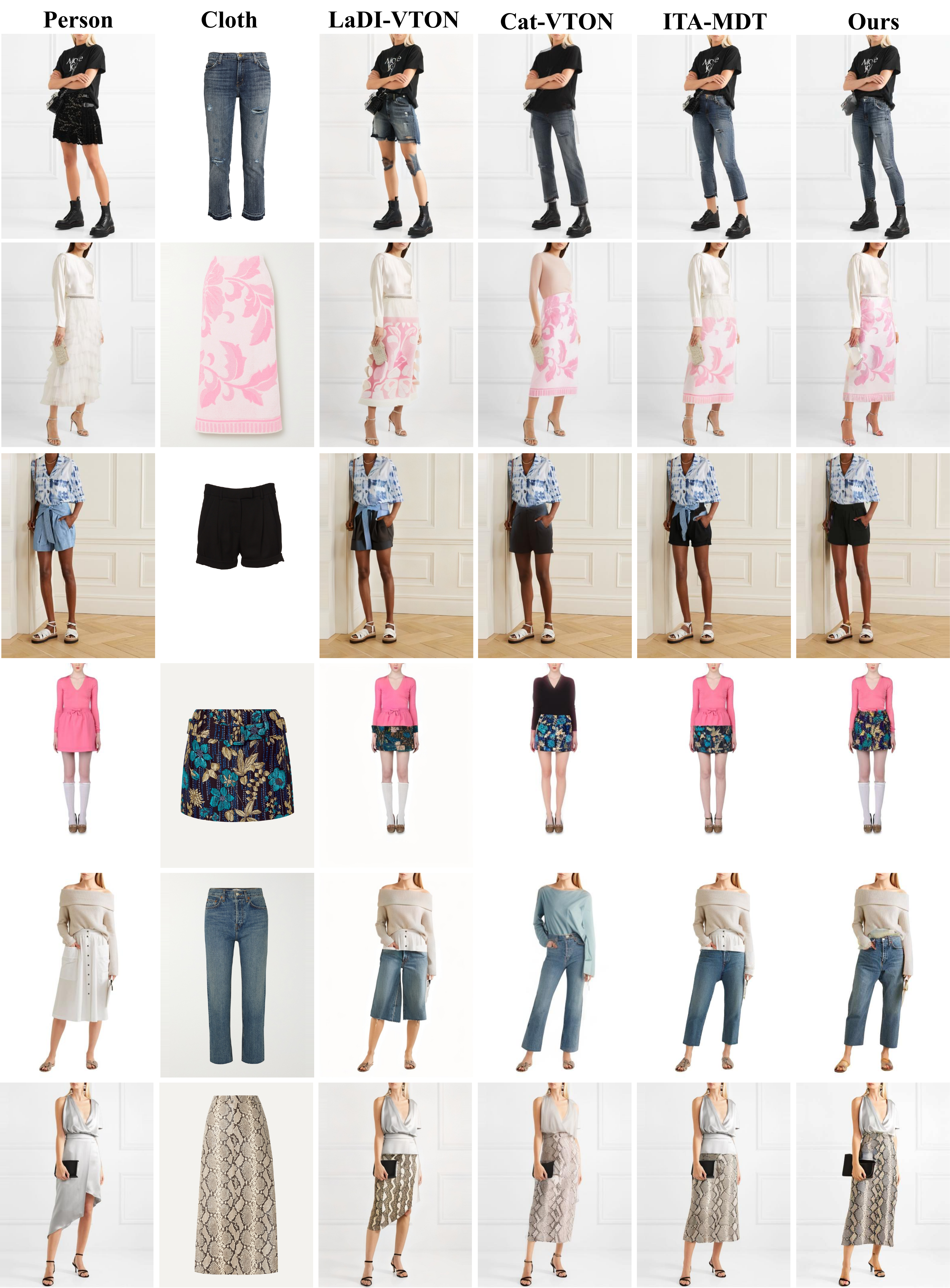}
   \captionsetup{labelfont=scriptsize,textfont=scriptsize}
   \caption{Qualitative results on DressCode \cite{morelli2022dresscode} Lower. Our model (Ours) here is trained only with VITON-HD \cite{choi2021viton} and DressCode \cite{morelli2022dresscode}.}
   \label{fig:results-dresscode-lower}
\end{figure*}

\begin{figure*}[t]
  \centering
   \includegraphics[width=0.9\linewidth]{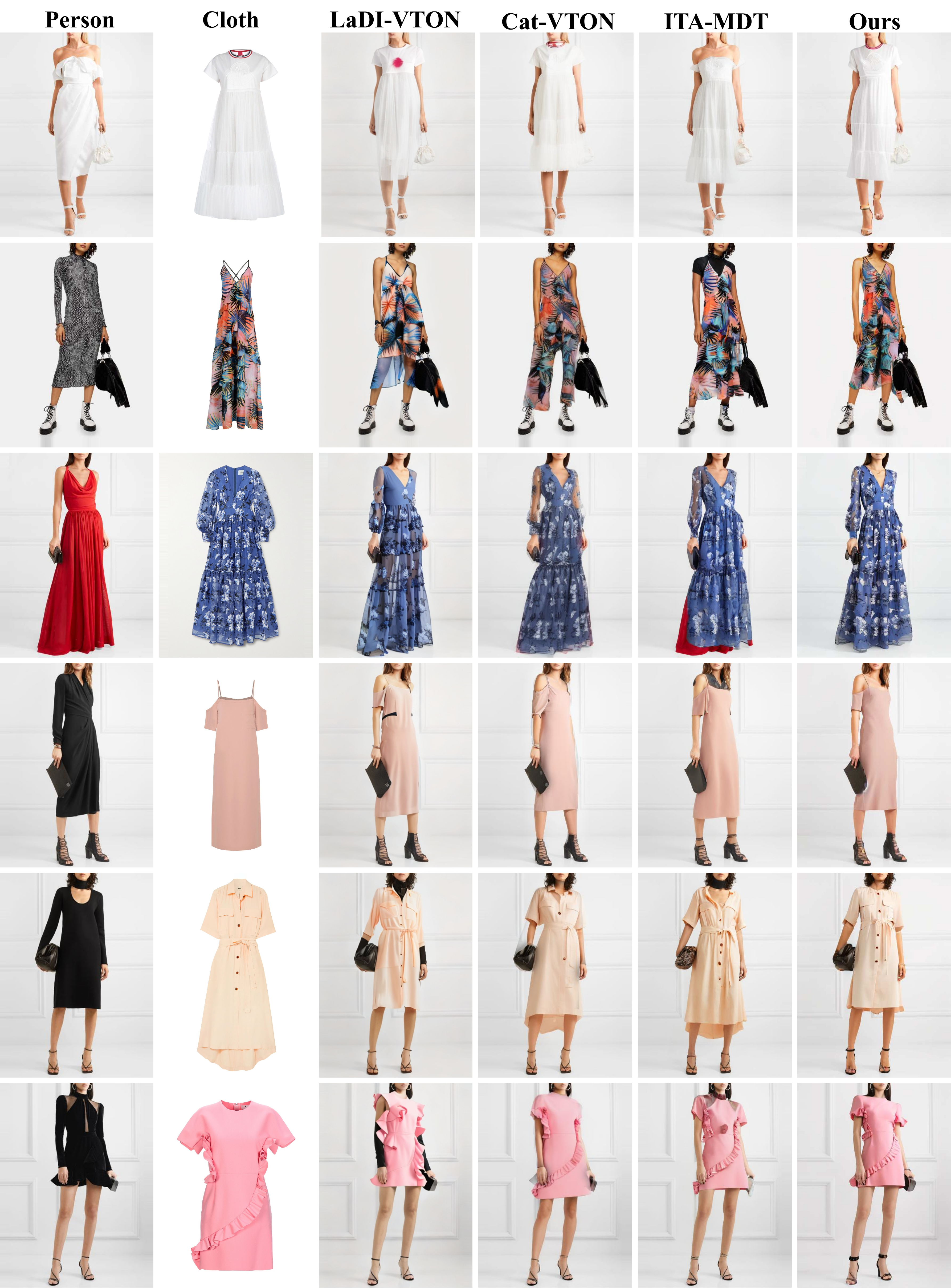}
   \captionsetup{labelfont=scriptsize,textfont=scriptsize}
   \caption{Qualitative results on DressCode \cite{morelli2022dresscode} Dresses. Our model (Ours) here is trained only with VITON-HD \cite{choi2021viton} and DressCode \cite{morelli2022dresscode}.}
   \label{fig:results-dresscode-dresses}
\end{figure*}

This section presents additional results on the DressCode benchmark. Tab.~\ref{tab:fullresult-dresscode} extends Tab.~\ref{tab:fullresult} in the main paper by reporting results on the Lower and Dresses subsets. FlexTry-On achieves the best unpaired FID on Lower and the second-best on Dresses. All baselines are evaluated using their official checkpoints under identical settings. Qualitative comparisons are shown in  Fig.~\ref{fig:results-dresscode-lower}, and Fig.~\ref{fig:results-dresscode-dresses}.

For lower-body try-on, our approach generates higher-fidelity outputs and better preserves surrounding regions such as shoes, socks, and upper garments (examples 1--5 in Fig.~\ref{fig:results-dresscode-lower}). In comparison, CatVTON still alters clothing regions, while ITA-MDT struggles to maintain consistent structure and realism in the preserved areas.

For dress try-on, prior methods often misplace the dress or alter its style, and may produce artifacts around the neck and hands (examples 1, 2, 4, 5, and 6 in Fig.~\ref{fig:results-dresscode-dresses}). In contrast, our method maintains more accurate garment structure and appearance while reducing such artifacts.

\section{Cross-Domain Qualitative Results}
\label{sec:Crossdomain_discussion}
This section presents additional qualitative results on cross-domain virtual try-on, where the garment category of the reference input differs from that of the target person. Such settings are particularly challenging because they violate the category-consistency assumption commonly adopted by existing virtual try-on methods.

\begin{figure*}[t]
  \centering
   \includegraphics[width=0.92\linewidth]{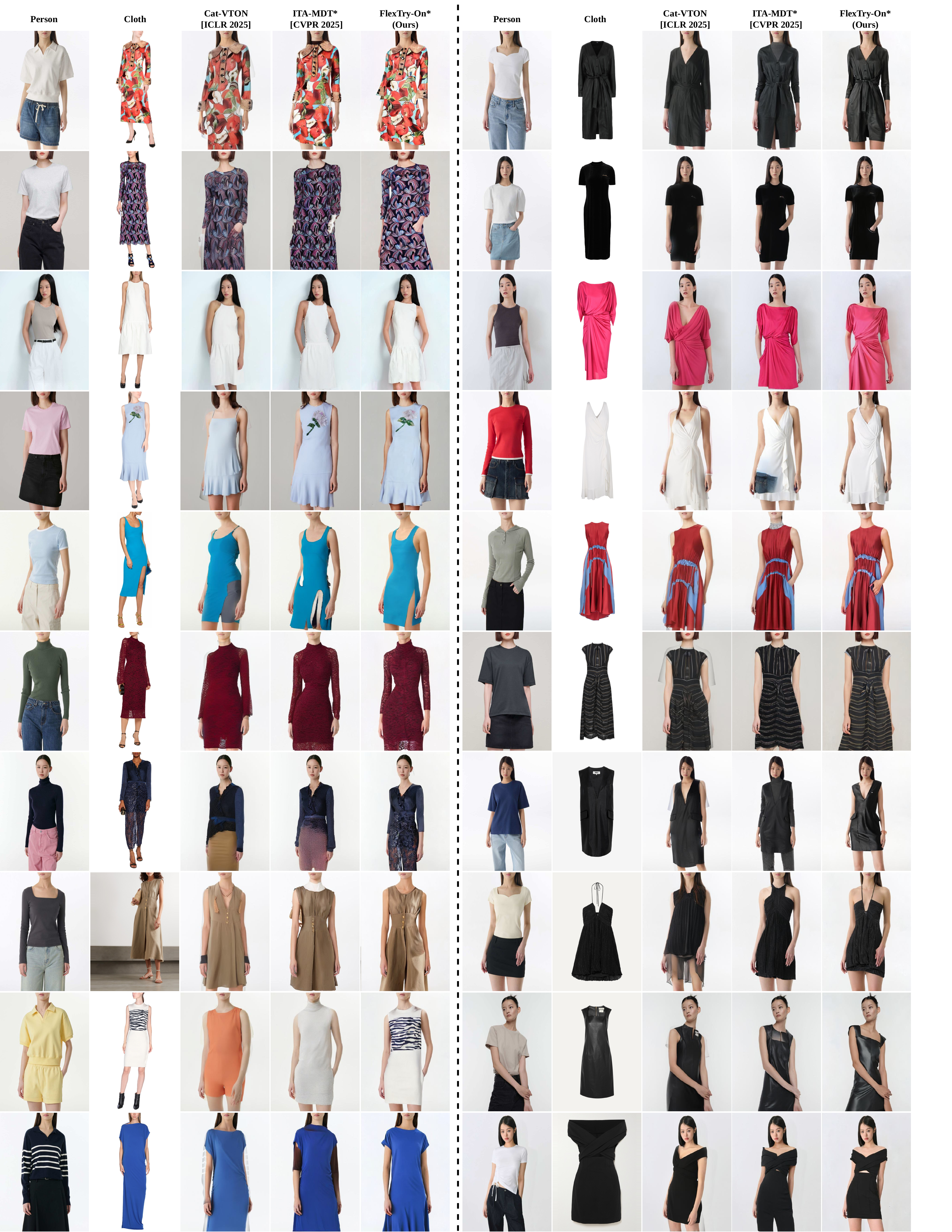}
   \captionsetup{labelfont=scriptsize,textfont=scriptsize}
   \caption{Qualitative results on Cross Domain (Upper-Lower to Dress) with C2P and P2P Setting. The dresses are taken from DressCode \cite{morelli2022dresscode}, and the person's images are taken from the Internet. FlexTry-On* refers to our training on the combined uncurated datasets. ITA-MDT* \cite{hong2025ita} denotes the fine-tuned ITA-MDT with segmented clothing, trained on our dataset. }
   \label{fig:results-cross-cloth-person-Dress}
\end{figure*}

One representative scenario involves transferring a dress or full-body garment onto a person wearing separate upper and lower garments. As shown in Fig.~\ref{fig:results-cross-cloth-person-Dress}, FlexTry-On produces realistic results in both C2P and P2P settings while preserving fine-grained garment details, including complex patterns, transparency effects, and structural characteristics.

A second scenario considers transferring upper or lower garments onto a person wearing a dress. This requires determining which regions should be preserved and which should participate in the try-on process. By leveraging text-aware garment masks, FlexTry-On can effectively handle these ambiguous boundaries and generate coherent results across different category transformations. Examples are shown in Fig.~\ref{fig:results-cross-cloth-person-Upper}, Fig.~\ref{fig:results-cross-person-person-Upper}, Fig.~\ref{fig:results-cross-cloth-person-Lower}, and Fig.~\ref{fig:results-cross-person-person-Lower}.

In contrast, methods relying on conventional human parsing~\cite{li2020self}
do not natively define task-specific editable regions for cross-domain
configurations, particularly dress-to-upper and dress-to-lower try-on.
For visualization purposes, when evaluating ITA-MDT* and CatVTON in these
settings, we manually mask all clothing regions on the input person to define
the inpainting area. Even with this manually adapted masking strategy,
FlexTry-On produces more coherent and visually consistent results. These
examples demonstrate how the proposed unified framework extends beyond
conventional C2P and P2P settings to challenging cross-domain virtual try-on
scenarios.

\section{Text-aware Segmentation Module Training Configuration}
\label{sec:tasm_training_config}


In this section, we provide the handcrafted text conditions used to train TASM and discuss the rationale behind their design. Unlike conventional human parsing, which assigns each pixel to a fixed semantic category, TASM aims to produce task-oriented masks that can support diverse virtual try-on scenarios, including cross-domain settings. Therefore, the predefined text conditions are organized according to functional garment regions rather than dataset-specific semantic labels.

The classes and corresponding text conditions are listed below:

\begin{enumerate}
\item Upper body Clothes: \textit{'inner shirt or inner t-shirt or top, or bra'}
\item Lower body Clothes: \textit{'shorts, or pants, or skirt, or trousers, or underwear'}
\item Full body Clothes: \textit{'dresses, or jumpsuits, or overalls'}
\item Outer Clothes: \textit{'jackets, or blazers, or outer shirt, or long padding jackets'}
\item Face and hand: \textit{'face and hand'}
\item Shoes: \textit{'shoes'}
\item Bag: \textit{'bags'}
\item Upper Body Skin (neck and arms): \textit{'upper body skin'}
\item Lower Body Skin (legs): \textit{'lower body skin'}
\item Full Body Skin (neck, arms, and legs): \textit{'full body skin'}
\end{enumerate}

The region-preservation strategy is designed according to the target garment category. Intuitively, regions unrelated to the target garment should be preserved whenever possible to maintain visual consistency. For example, during upper-body try-on, lower garments and lower-body skin are retained, whereas during lower-body try-on, upper garments and upper-body skin are preserved. For full-body and outerwear try-on, only identity-related and accessory regions, such as faces, hands, shoes, and bags, are retained.

This design is particularly important for cross-domain virtual try-on. Consider the case of transferring an upper garment onto a person wearing a dress. Conventional human parsing typically treats the entire dress as a single semantic region, making it difficult to determine which parts should be preserved and which should participate in the try-on process. In contrast, the proposed text-conditioned decomposition enables TASM to generate task-specific masks, providing more flexible region control across diverse garment categories and cross-domain scenarios.

\vspace{-3mm}
\section{Implementation Details}
\label{sec:imp_detail}
This section provides additional implementation details of the proposed framework. For preprocessing, we use MediaPipe~\cite{moryossef2024optimizing} for pose keypoint extraction and DensePose~\cite{guler2018densepose} for dense body correspondence estimation.

The proposed Dual-DiT Tryon Module (DDTM) is built upon DiT-XL/2~\cite{peebles2023scalable}, with 28 transformer layers, including 4 decoding layers, a hidden dimension of 1152, and 16 attention heads. All images are resized to $512\times512$. Following prior work, we employ the SD 2.0 VAE~\cite{rombach2022high}, which encodes images into latent representations of size $64\times64\times4$.

During training, we use 1000 diffusion timesteps and perform inference with 30 denoising steps. The model is optimized using a learning rate of $1\times10^{-4}$ and an EMA decay of 0.9999. All remaining optimization settings, including the optimizer and learning-rate scheduler, follow the original DiT configuration~\cite{peebles2023scalable}. The classifier-free guidance scale is set to 2.0. Unless otherwise specified, reported results correspond to the model trained for 1.5M iterations.

Training is conducted on either 8 NVIDIA A6000 GPUs (48GB each) or 4 NVIDIA H200 GPUs (141GB each), depending on hardware availability. Within every ablation pair, the variants with and without $\mathcal{L}_{style}$ use the same batch size, optimizer, learning-rate schedule, and number of training iterations; only the loss term is changed.

The Text-Aware Segmentation Module (TASM) is built upon DiT-B/2, consisting of 12 transformer layers, a hidden dimension of 768, and 12 attention heads. TASM is trained for 800K iterations on a single RTX 4090 GPU (24GB) with a batch size of 7.

For inference-time measurement, all methods are benchmarked on an RTX 4090 GPU with 24GB memory whenever possible. Since OmniTry-On and Any2Any Try-On cannot be directly evaluated on this platform, we first measure their runtime on an RTX A6000 GPU with 48GB memory and estimate the corresponding RTX 4090 runtime using the runtime scaling factor of our method measured on both GPUs. The controlled module comparison reports online segmentation and denoising; pose, DensePose, and other preprocessing are identical across the compared variants and therefore excluded.

\resubmit{For the G-SAM baseline~\cite{liu2023grounding,ren2024grounded}, we use GroundingDINO-Tiny followed by SAM-B with their publicly released checkpoints, without any additional fine-tuning. Runtime is measured for one batch-size-1 try-on. Each segmentation method must evaluate all 10 segmentation-class conditions listed above for the same input image. TASM evaluates these conditions together as one batch in 0.1 s. In contrast, the GroundingDINO--SAM pipeline processes the 10 class conditions sequentially, requiring approximately 1.9 s per class, or 19.0 s per try-on. The subsequent DDTM denoising stage takes 1.1 s per try-on for either segmentation method.}


\begin{table}[t]
\centering
\captionsetup{labelfont=scriptsize,textfont=scriptsize}
\caption{IVTON Data Statistic
}
\vspace{-3mm}
\label{tab:data_tab}
\scalebox{0.8}{
\begin{tabular}{l|cccc}
\multicolumn{1}{c|}{Dataset}  & Resolutions & C2P & P2P & \# of images               \\ 
\hline
VITON-HD  \cite{choi2021viton}    & 1024 x 768  & \textcolor{green}{\ding{51}} & \textcolor{red}{\ding{55}}                    & 11 647                     \\
DressCode \cite{morelli2022dresscode}    & 1024 x 768  &\textcolor{green}{\ding{51}} & \textcolor{red}{\ding{55}}                      & 53 792                     \\
DeepFashion \cite{liu2016deepfashion}   & Varies      & \textcolor{red}{\ding{55}}  & \textcolor{green}{\ding{51}}                     & 52 712                     \\
IMAGDress   \cite{shen2025imagdressing}  & Varies      & \textcolor{green}{\ding{51}} & \textcolor{green}{\ding{51}}                     & 324 857                    \\
LRVS-Fashion \cite{lepage2023lrvsf}& Varies      & \textcolor{green}{\ding{51}} & \textcolor{green}{\ding{51}}                     & 500 000  \\
\hline
\end{tabular}}
\vspace{-6mm}
\end{table}
\section{Dataset and Evaluation}
\label{sec:dataset}

This section describes the datasets used for training and evaluation.
To support diverse virtual try-on scenarios, we construct a unified training
corpus from VITON-HD~\cite{choi2021viton},
DressCode~\cite{morelli2022dresscode},
DeepFashion~\cite{liu2016deepfashion},
IMAGDress~\cite{shen2025imagdressing}, and a filtered subset of
LRVS-Fashion~\cite{lepage2023lrvsf}. These datasets provide both C2P and
P2P samples across diverse garment categories, poses, identities, and
backgrounds.

VITON-HD provides 11,647 training and 2,032 test images at
$1024\times768$. DressCode contains over 50,000 images covering upper-body,
lower-body, and dresses. DeepFashion contributes 52,712 images for P2P
training. IMAGDress contains 324,857 paired samples, while approximately
500,000 downloadable LRVS-Fashion images are retained after filtering.
All datasets are standardized into four garment categories, yielding
approximately 930,000 training samples in total. Detailed statistics are
reported in Tab.~\ref{tab:data_tab}. We report LPIPS~\cite{zhang2018unreasonable},
SSIM~\cite{wang2004image}, and FID~\cite{heusel2017gans}.
Lower LPIPS/FID and higher SSIM indicate better performance. \resubmit{To evaluate inference-seed sensitivity, we repeat inference on VITON-HD with seeds $\{0,1,2\}$ using the same checkpoint. Seed-wise paired/unpaired FID is reported in Tab.~\ref{tab:fid_seeds}(b).}

We report both paired and unpaired FID. Paired FID evaluates reconstruction
against corresponding ground truth, whereas unpaired FID evaluates unseen
garment--person combinations and serves as the primary metric for the
standard unpaired IVTON setting.



\section{Discussion}
\label{sec:suppl_discussion}

\subsection{Inpainting Mask Strategy}
\label{sec:inpainting_mask_strategy}

In most prior IVTON methods, the inpainting mask is determined by the
try-on garment category. For example, when the target garment is an
upper-body item (e.g., a T-shirt or shirt), the mask is typically constructed
from the corresponding upper-body garment region of the person together with
exposed skin regions, while preserving the hands, face, and hair. A similar
strategy is applied to lower-body garments. For full-body garments such as
dresses, the editable region generally covers all clothing regions and relevant
body areas while preserving identity-related regions such as the face, hair,
and hands.

However, such category-dependent masking becomes ambiguous in cross-domain
try-on settings. For instance, when a person is wearing a dress, conventional
human parsing does not directly determine which part of the dress should be
edited when the target garment is an upper- or lower-body item. Consequently,
existing category-based masking strategies cannot naturally define the
corresponding editable region. TASM addresses this issue by producing
task-aware, text-conditioned masks according to the target garment category.

\begin{figure*}[t]
  \centering
   \includegraphics[width=0.88\linewidth]{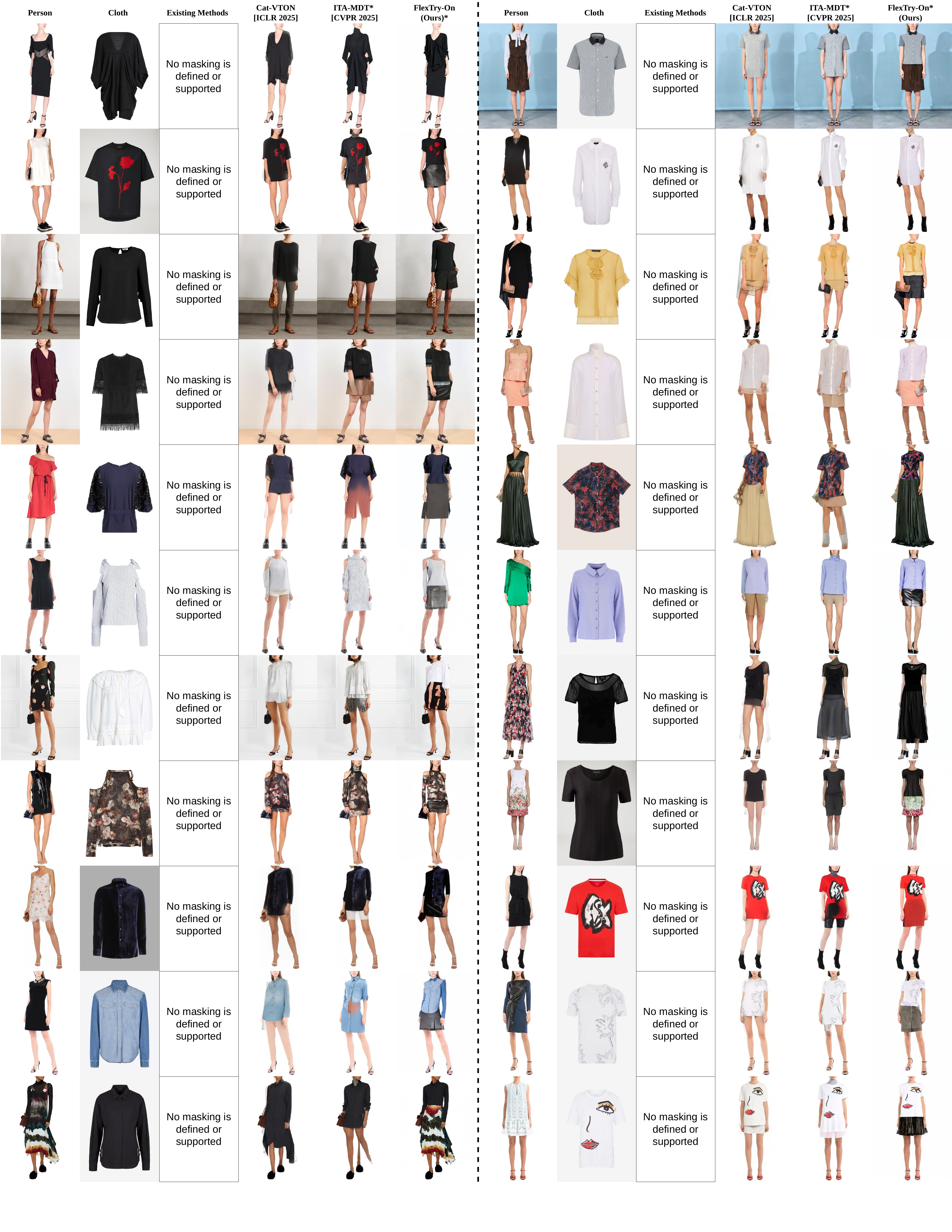}
   \captionsetup{labelfont=scriptsize,textfont=scriptsize}
   \caption{Qualitative results on Cross Domain (Dress to Upper) with C2P Setting. The Upper Garment and person's image are taken from DressCode \cite{morelli2022dresscode}. ITA-MDT* \cite{hong2025ita} denotes the fine-tuned ITA-MDT with segmented clothing, trained on our dataset. FlexTry-On* refers to training on the combined uncurated datasets. Existing methods \cite{chong2024catvton,choi2024improving,hong2025ita,han2018viton,dam2024time,kim2024stableviton,kim2025promptdresser,morelli2023ladi} based on conventional human parsing \cite{li2020self} are limited in supporting flexible masking for cross-domain try-on. In this experiment, we either masked all clothing or used full-body clothing masks for ITA-MDT and CatVTON.}
   \label{fig:results-cross-cloth-person-Upper}
\end{figure*}

\begin{figure*}[t]
  \centering
   \includegraphics[width=1.0\linewidth]{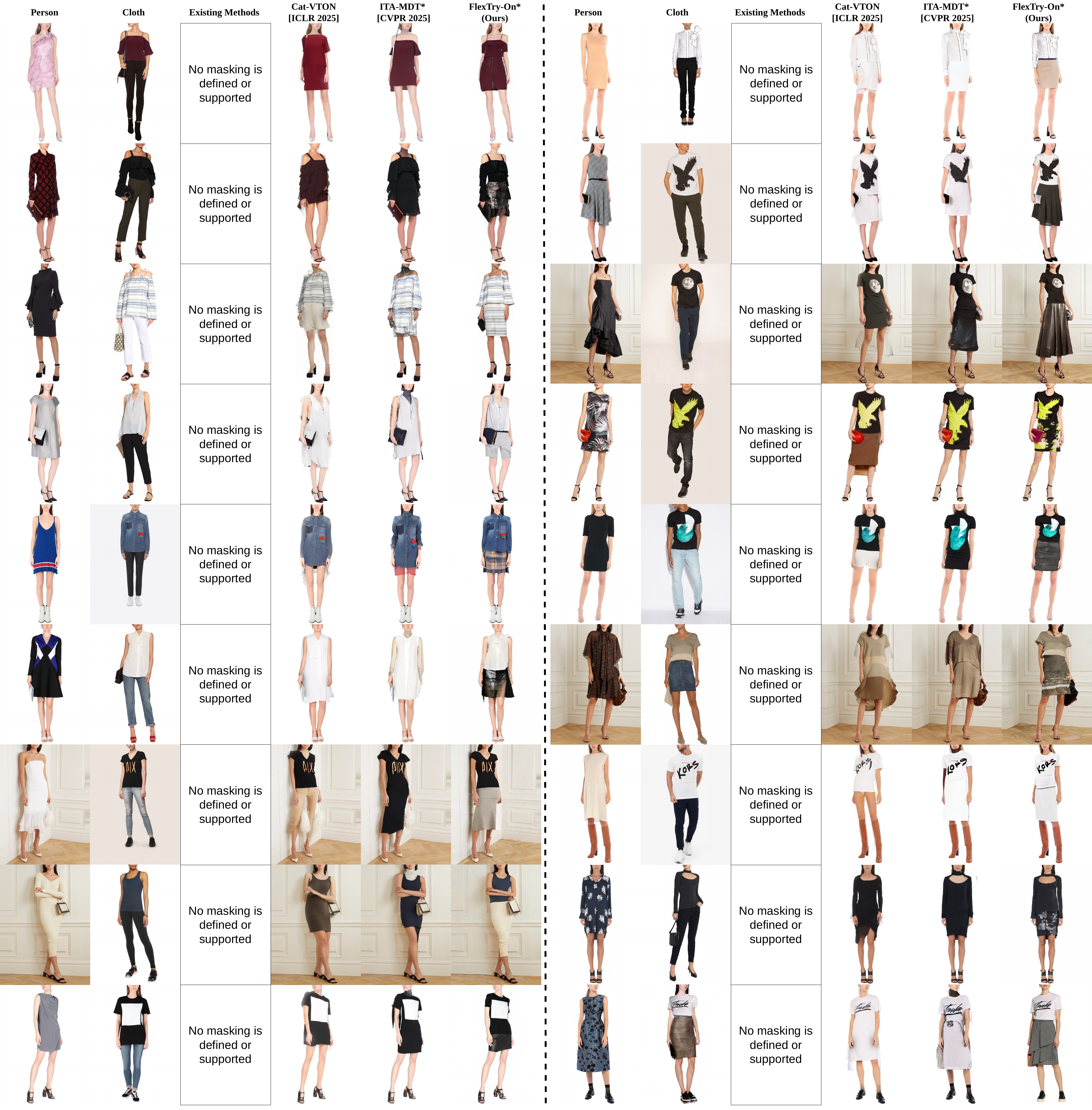}
   \captionsetup{labelfont=scriptsize,textfont=scriptsize}
   \caption{Qualitative results on Cross Domain (Dress to Upper) with P2P Setting. The Upper Garment and person's image are taken from DressCode \cite{morelli2022dresscode}. ITA-MDT* \cite{hong2025ita} denotes the fine-tuned ITA-MDT with segmented clothing, trained on our dataset. FlexTry-On* refers to training on the combined uncurated datasets. Existing methods \cite{chong2024catvton,choi2024improving,hong2025ita,han2018viton,dam2024time,kim2024stableviton,kim2025promptdresser,morelli2023ladi} typically rely on category-specific masks derived from conventional human parsing pipelines \cite{li2020self}. In cross-domain scenarios, however, the editable regions cannot be uniquely determined from predefined garment categories. For ITA-MDT and CatVTON, we therefore evaluate two representative masking strategies: masking all clothing regions and using full-body clothing masks. In contrast, TASM generates task-aware masks conditioned on the target garment description.}
   \label{fig:results-cross-person-person-Upper}
\end{figure*}

\begin{figure*}[t]
  \centering
   \includegraphics[width=0.9\linewidth]{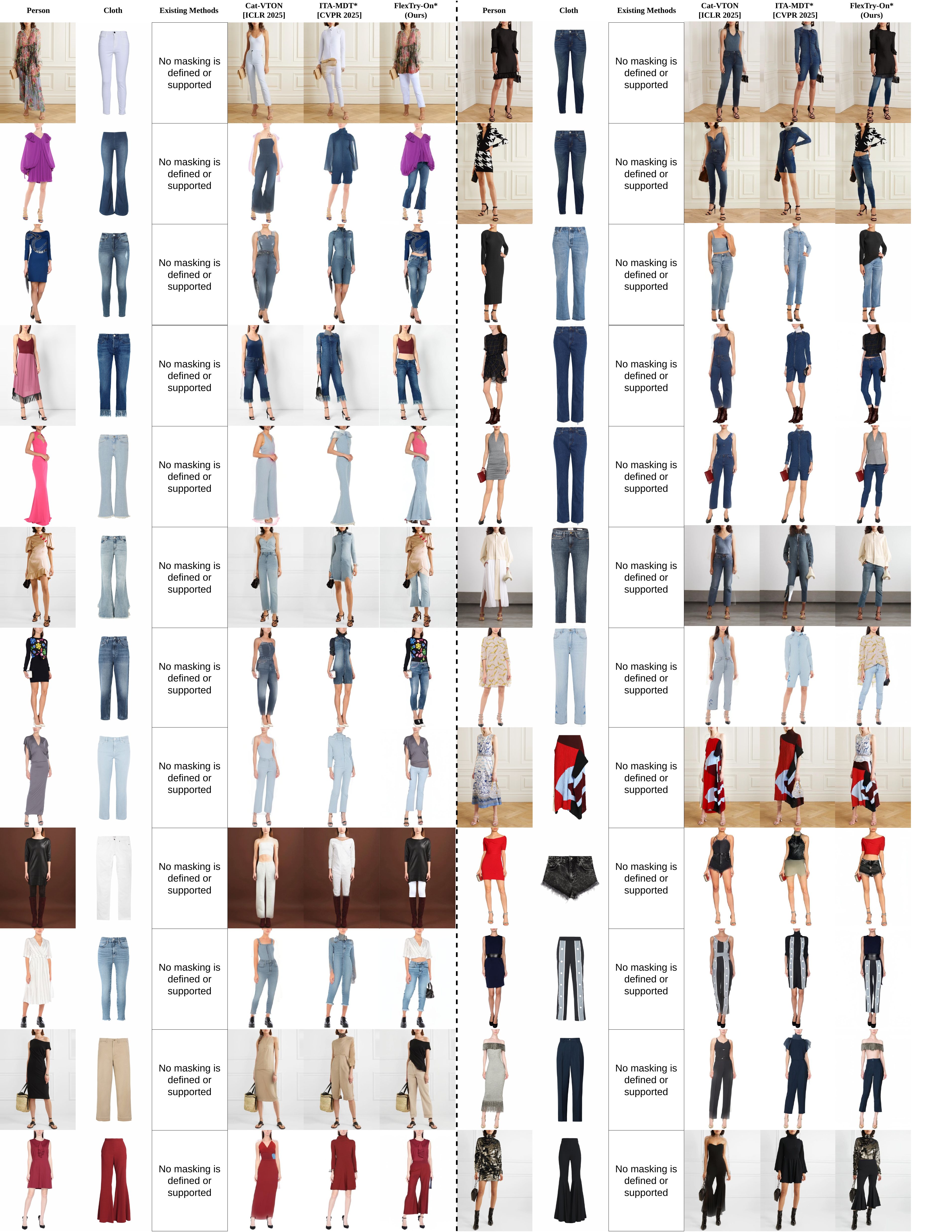}
   \captionsetup{labelfont=scriptsize,textfont=scriptsize}
   \caption{Qualitative results on Cross Domain (Dress to Lower) with C2P Setting. The Lower Garment and person's image are taken from DressCode \cite{morelli2022dresscode}. ITA-MDT* \cite{hong2025ita} denotes the fine-tuned ITA-MDT with segmented clothing, trained on our dataset. FlexTry-On* refers to training on the combined uncurated datasets. Existing methods \cite{chong2024catvton,choi2024improving,hong2025ita,han2018viton,dam2024time,kim2024stableviton,kim2025promptdresser,morelli2023ladi} typically rely on category-specific masks derived from conventional human parsing pipelines \cite{li2020self}. In cross-domain scenarios, however, the editable regions cannot be uniquely determined from predefined garment categories. For ITA-MDT and CatVTON, we therefore evaluate two representative masking strategies: masking all clothing regions and using full-body clothing masks. In contrast, TASM generates task-aware masks conditioned on the target garment description.
}
   \label{fig:results-cross-cloth-person-Lower}
\end{figure*}

\begin{figure*}[t]
  \centering
   \includegraphics[width=0.88\linewidth]{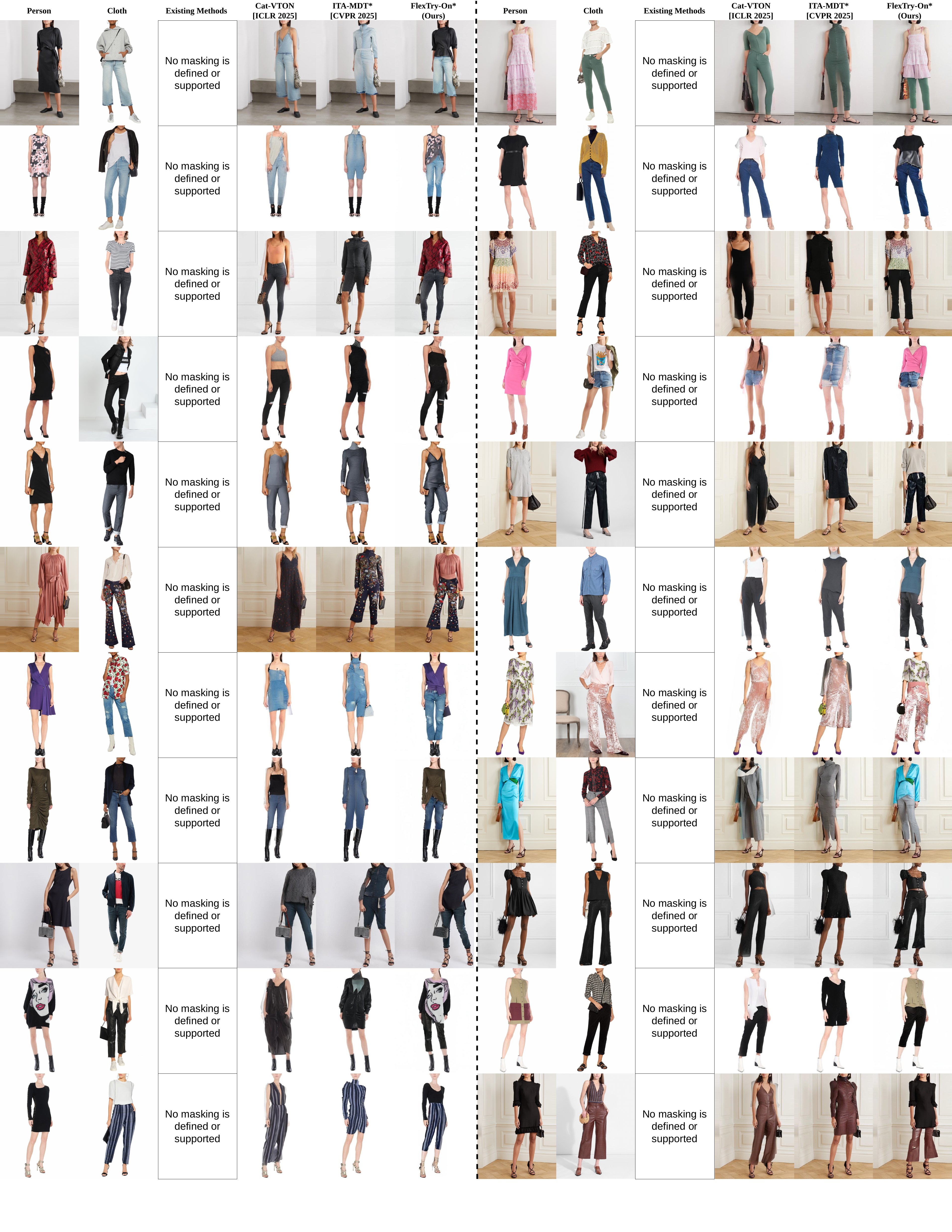}
   \captionsetup{labelfont=scriptsize,textfont=scriptsize}
   \caption{Qualitative results on Cross Domain (Dress to Lower) with P2P Setting. The Lower Garment and person's image are taken from DressCode \cite{morelli2022dresscode}. ITA-MDT* \cite{hong2025ita} denotes the fine-tuned ITA-MDT with segmented clothing, trained on our dataset. FlexTry-On* refers to training on the combined uncurated datasets. Existing methods \cite{chong2024catvton,choi2024improving,hong2025ita,han2018viton,dam2024time,kim2024stableviton,kim2025promptdresser,morelli2023ladi} typically rely on category-specific masks derived from conventional human parsing pipelines \cite{li2020self}. In cross-domain scenarios, however, the editable regions cannot be uniquely determined from predefined garment categories. For ITA-MDT and CatVTON, we therefore evaluate two representative masking strategies: masking all clothing regions and using full-body clothing masks. In contrast, TASM generates task-aware masks conditioned on the target garment description.
}
   \label{fig:results-cross-person-person-Lower}
\end{figure*}

For qualitative comparison in dress-to-upper and dress-to-lower settings, we
adapt prior methods by treating the editable clothing area as a full-body
region, allowing them to generate outputs under their existing masking
pipelines. The resulting comparisons are shown in
Fig.~\ref{fig:cross-domain} in the main paper and
Figs.~\ref{fig:results-cross-cloth-person-Upper},
\ref{fig:results-cross-person-person-Upper},
\ref{fig:results-cross-cloth-person-Lower}, and
\ref{fig:results-cross-person-person-Lower}.

\begin{figure}[t]
    \centering
   \includegraphics[width=1.0\linewidth]{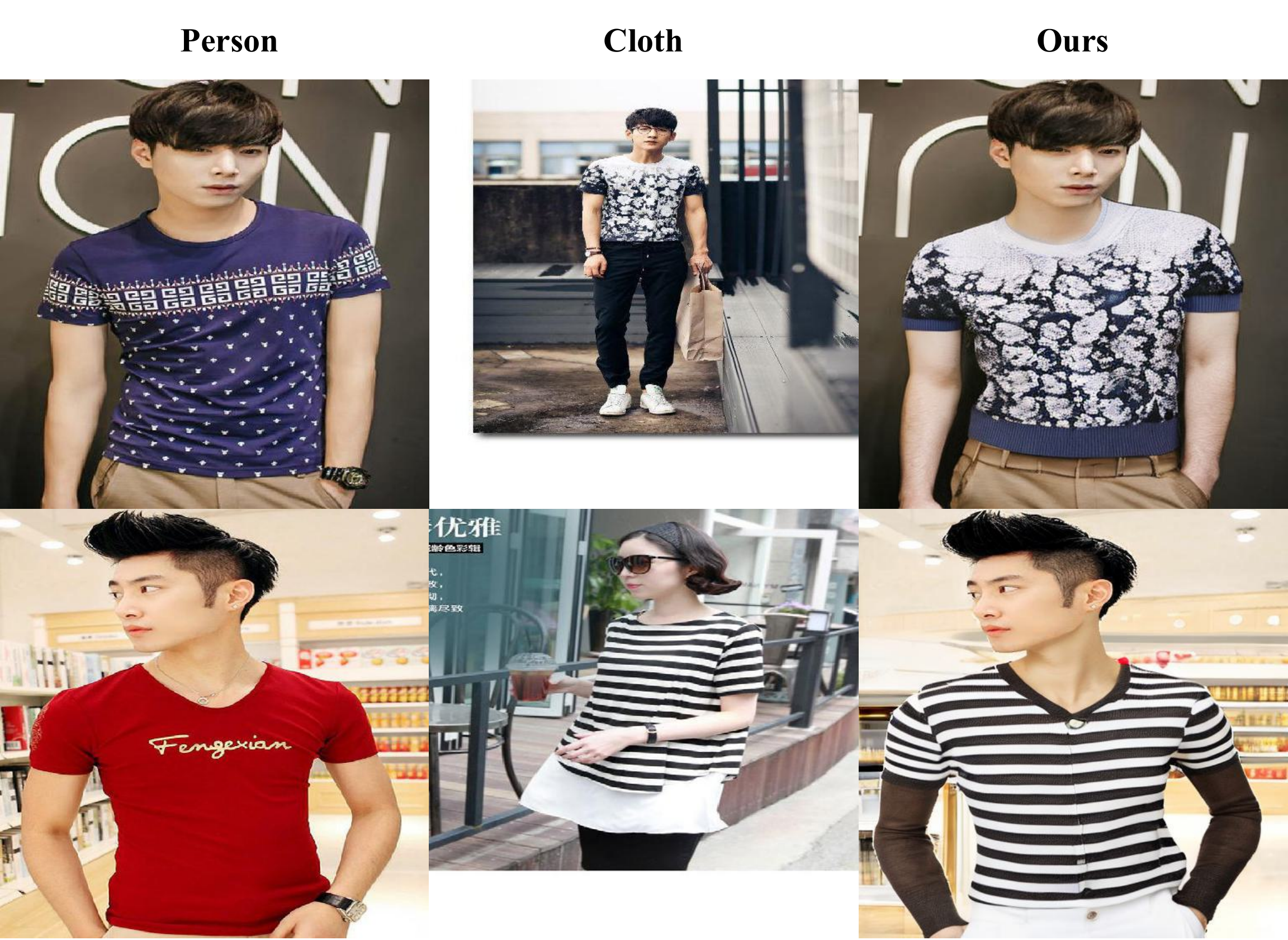}
   \vspace{-4mm}
   \captionsetup{labelfont=scriptsize,textfont=scriptsize}
   \caption{\resubmit{Representative Street2Street failure cases on Street Try-On~\cite{cui2025street}. FlexTry-On may preserve the garment's appearance while failing to fully reproduce its fine-grained structure.}}
   \label{fig:failure}
   \vspace{-2mm}
\end{figure}

\subsection{\resubmit{Evaluation on Diverse Subjects}}
\label{sec:diverse_subjects}

\begin{figure}[b]
    \centering
    \includegraphics[width=1.0\linewidth]{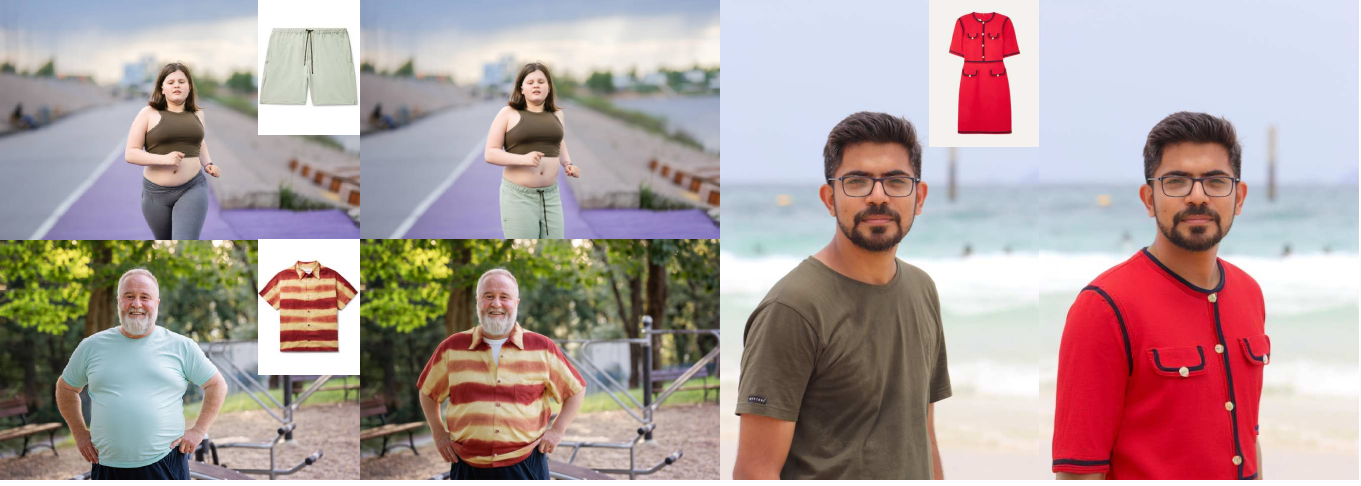}
    \captionsetup{labelfont=scriptsize,textfont=scriptsize}
    \caption{\resubmit{
    Additional qualitative results on heterogeneous subjects with varied
    appearances, body shapes, poses, framing, and backgrounds. All outputs
    are generated using the same FlexTry-On* checkpoint without additional
    adaptation.
    }}
    \label{fig:diverse_subjects}
    \vspace{0mm}
\end{figure}

\resubmit{
To further examine generalization beyond curated fashion-style examples,
we provide additional qualitative results on heterogeneous subjects with
varied appearances, body shapes, poses, framing, and background conditions
in Fig.~\ref{fig:diverse_subjects}. All examples are generated using the same
FlexTry-On* checkpoint without additional adaptation. These results broaden
the qualitative evaluation beyond conventional benchmark-style images,
although they do not establish demographic robustness.
}

\subsection{\resubmit{Evaluation Scope and Reproducibility}}

\resubmit{
Our quantitative evaluation covers standard C2P benchmarks and matched-data
C2P--P2P evaluation. Specifically, C2P is evaluated using paired and unpaired
VITON-HD and DressCode protocols, while Tab.~\ref{tab:fullresult-abl}
compares methods retrained on the same C2P--P2P corpus. Because no
standardized paired benchmark exists for the category-changing transitions
considered in our cross-domain experiments, we restrict this capability to
zero-shot qualitative evaluation.
}

\resubmit{
Paired metrics evaluate reconstruction against corresponding targets, whereas
unpaired FID evaluates unseen garment--person combinations and remains the
primary metric for standard unpaired IVTON. Using the same checkpoint, three
inference seeds yield paired/unpaired FID standard deviations of 0.02/0.01,
indicating low sensitivity to sampling randomness.
}

\subsection{\resubmit{Efficiency and Computational Footprint}}

\resubmit{The controlled TASM--G-SAM comparison isolates segmentation as the only
differing component. TASM evaluates all 10 text conditions jointly in 0.1~s,
whereas G-SAM processes them sequentially in approximately 19.0~s. Both are
followed by the same 1.1~s DDTM inference. Shared pose and DensePose
preprocessing is excluded from this controlled comparison. Precomputing
G-SAM masks would remove this online segmentation cost, but would not affect
the downstream conditioning-quality comparison in Tab.~\ref{tab:abl_main}.}

\resubmit{FlexTry-On contains 1.634B parameters across its three DiT modules and
2.058B parameters when including the frozen VAE and CLIP text encoder.
At $512\times512$ resolution, batch size 1, and BF16 precision, inference
requires 5.15~GB peak allocated GPU memory.}

\section{Limitations}
\label{sec:suppl_limitations}

Our framework is built upon an inpainting-based formulation and therefore
inherits limitations common to diffusion-based virtual try-on methods.
In particular, extremely oversized or unconventional garments remain
challenging when their desired geometry extends substantially beyond the
editable region.

\resubmit{Performance may also degrade under severe occlusion or for highly complex
garment structures that are underrepresented in the training data.
Street-to-Street transfer is particularly challenging because both the source
garment and target person may exhibit uncontrolled pose, illumination,
background, garment appearance, and occlusion. Moreover, when the reference
garment occupies only a small region of the source image, fine-grained garment
cues become less informative, making garment-identity preservation more
difficult and increasing the tendency to hallucinate garment details.}

\resubmit{Representative failure cases are shown in Fig.~\ref{fig:failure}. In these
examples, FlexTry-On captures the overall garment appearance but may fail to
faithfully preserve fine-grained details and geometry, such as patterns,
necklines, sleeve shapes, and silhouettes. These failures are more pronounced
when the reference garment is small or partially occluded, suggesting that
limited garment evidence remains a key challenge in unconstrained
Street-to-Street settings.}

\resubmit{The current preprocessing pipeline assumes that a single visible target person
has been selected before inference. It does not automatically handle multiple
people, person-to-person occlusion, or ambiguous target selection in crowded
scenes. In addition, the combined training corpus inherits demographic and
social-media collection biases; our current evaluation does not establish
uniform performance across demographic groups.}

\resubmit{These limitations motivate future work on more robust garment representations
under limited visual evidence, automatic multi-person target selection, and
broader, demographically balanced virtual try-on datasets.}


\end{document}